\documentclass[runningheads]{llncs}

\usepackage[mobile]{eccv}

\usepackage{eccvabbrv}

\usepackage{graphicx}
\usepackage{booktabs}

\usepackage[accsupp]{axessibility}  % Improves PDF readability for those with disabilities.

\usepackage[pagebackref,breaklinks,colorlinks,citecolor=eccvblue]{hyperref}
\usepackage{orcidlink}
\usepackage{marvosym}
\usepackage{colortbl}
\usepackage{multirow}
\usepackage{rotating}
\usepackage{arydshln}
\definecolor{col1}{RGB}{232, 161, 148}
\definecolor{col2}{RGB}{148, 187, 232}
\usepackage{overpic}
\usepackage{xcolor}
\usepackage{float}
\usepackage{appendix}
\begin{document}

\title{Mono-ViFI: A Unified Learning Framework for Self-supervised Single- and Multi-frame Monocular Depth Estimation} 
% \title{Mono-ViFI: Supplementary Material} 

\titlerunning{Mono-ViFI: A Unified Framework for Self-supervised Monocular Depth}

\author{Jinfeng Liu \and
Lingtong Kong\textsuperscript{\Letter} \and
Bo Li \and Zerong Wang \and Hong Gu \and Jinwei Chen}

\authorrunning{J.~Liu et al.}

\institute{vivo Mobile Communication Co., Ltd, China\\
\email{\{liujinfeng,ltkong,libra,wangzerong,guhong,jinwei.chen\}@vivo.com}}

\maketitle

\begin{abstract}
% Self-supervised monocular depth estimation has gathered notable interest since it can liberate training from the dependency on depth annotations, and recent methods have achieved promising results. However, the latent information residing in the training data has yet to be thoroughly explored, which hinders the depth models at hand from realizing their full potential. To this end, we propose xxx, a generic and effective learning framework which focuses on monocular video training. To fully leverage existing video data, we adopt an affine augmentation and use an efficient video frame interpolation (VFI) model to generate more training instances. Moreover, we design a VFI-assisted multi-frame fusion depth model based on the single-frame depth network, which aggregates the temporal information for reasoning in virtue of the VFI model. Finally, we also incorporate self-distillation from data augmentation and mutual distillation between the single-frame model and multi-frame model, to further improve the performance. Extensive experiments on KITTI, Cityscapes and Make3D datasets demonstrate that xxx can bring significant improvements for current advanced depth models.
Self-supervised monocular depth estimation has gathered notable interest since it can liberate training from dependency on depth annotations. In monocular video training case, recent methods only conduct view synthesis between existing camera views, leading to insufficient guidance. To tackle this, we try to synthesize more virtual camera views by flow-based video frame interpolation (VFI), termed as temporal augmentation. For multi-frame inference, to sidestep the problem of dynamic objects encountered by explicit geometry-based methods like ManyDepth, we return to the feature fusion paradigm and design a VFI-assisted multi-frame fusion module to align and aggregate multi-frame features, using motion and occlusion information obtained by the flow-based VFI model. Finally, we construct a unified self-supervised learning framework, named Mono-ViFI, to bilaterally connect single- and multi-frame depth. In this framework, spatial data augmentation through image affine transformation is incorporated for data diversity, along with a triplet depth consistency loss for regularization. The single- and multi-frame models can share weights, making our framework compact and memory-efficient. Extensive experiments demonstrate that our method can bring significant improvements to current advanced architectures. Source code is available at \href{https://github.com/LiuJF1226/Mono-ViFI}{\textcolor{magenta}{https://github.com/LiuJF1226/Mono-ViFI}}.


\end{abstract}    
\section{Introduction}
\label{sec:intro}
Monocular depth estimation (MDE) is a fundamental 3D vision task, which finds applications in robotics, autonomous driving and VR/AR. To reduce the dependence on depth annotations, self-supervised MDE~\cite{garg,sfmlearner,godard,Uncertainty,deepdig,GasMono,dualrefine} has risen as a convenient and practical solution, which uses monocular videos, stereo pairs, or even both to generate supervision through view synthesis, where an image photometric reconstruction loss is minimized. Compared to stereo training case~\cite{garg,godard,hint} with known camera pose, learning from monocular videos~\cite{sfmlearner,mono2,manydepth} is more challenging since it requires an extra pose network to estimate camera ego-motion. Even so, it is much easier to collect monocular videos compared to stereo image pairs, making monocular training more attractive. Therefore, this paper also focus on self-supervised training with unlabeled monocular videos. 

\begin{figure*}[t]
\centering
\begin{overpic}[scale=0.2]{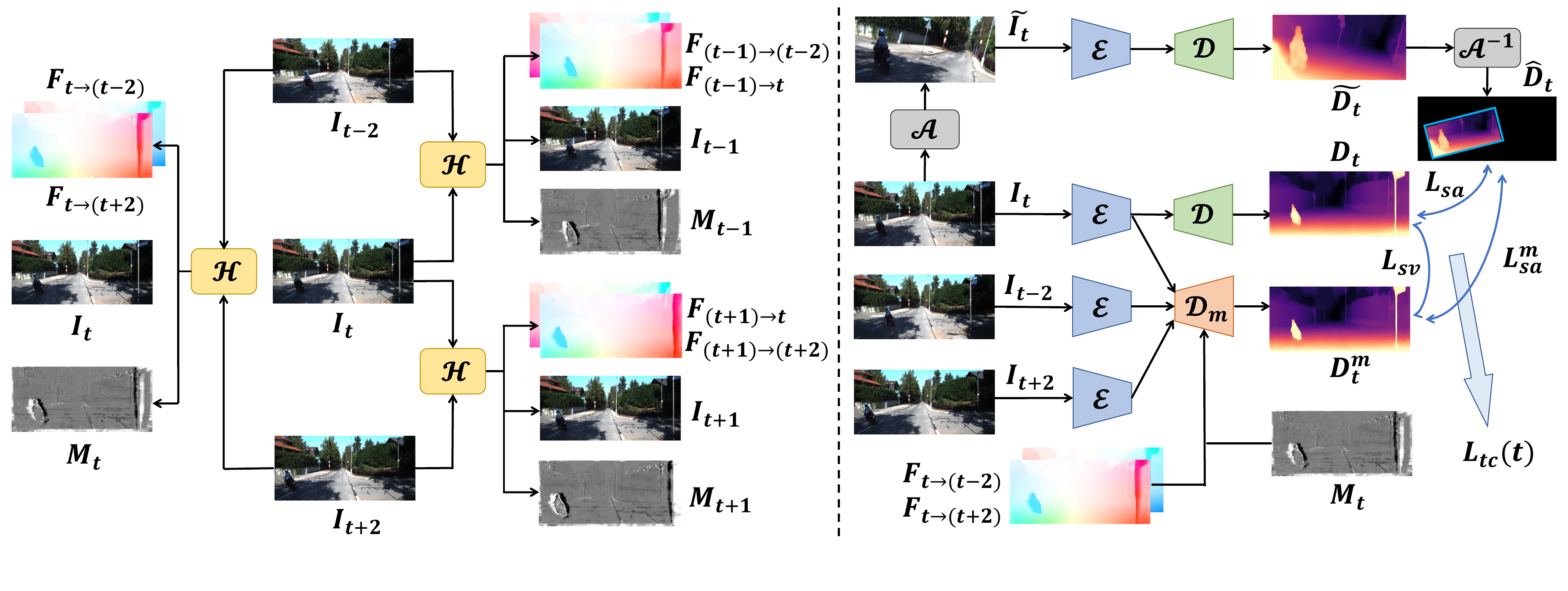}
\put(8,0){\scriptsize (a) Temporal Augmentation with VFI}
\put(55,0){\scriptsize (b) Learning Framework ($I_t$ for example)}
\end{overpic}
\caption{Overview of Mono-ViFI. (a) We achieve temporal augmentation by a flow-based VFI model $\mathcal{H}$, which also reasons out intermediate optical flow and occlusion mask for multi-frame inference. (b) For target image $I_t$, we obtain its single-frame depth $D_t$ and multi-frame depth $D_t^m$. Also, $I_t$ is augmented to $\widetilde{I}_t$ through affine transformation $\mathcal{A}$. We then calculate the depth $\widetilde{D}_t$ of $\widetilde{I}_t$ and inversely convert it to the original view, generating $\widehat{D}_t$. Finally, we enforce a triplet depth consistency loss $L_{tc}(t)$ among the three depth maps, including a standard view depth consistency loss $L_{sv}(t)$ and two scale-aware depth consistency losses, $L_{sa}(t)$ and $L_{sa}^{m}(t)$. Note that each depth map also corresponds to a photometric loss
and a smoothness loss, which are omitted here.}
\vspace{-2mm}
\label{fig:overview}
\end{figure*}

Recent works have made various attempts to improve accuracy, like introducing masking techniques~\cite{sfmlearner,mono2} and leveraging segmentation priors~\cite{sdo,iapc}. However, they only conduct view synthesis between existing fixed camera views within video datasets, leading to insufficient guidance especially in invisible regions by occlusion or disappearance. Considering that video data possesses temporal continuity, an intuitive idea is to generate novel frames, \ie, virtual camera views in the temporal dimension. We term it as temporal augmentation, which can be achieved by video frame interpolation (VFI), as shown in \cref{fig:overview}(a). 

From another perspective, multi-frame reasoning paradigm have recently been researched because multiple frames can be available for reasoning in many application scenarios. ManyDepth~\cite{manydepth} proposes to use cost volume constructed from multi-frame matching features. It explicitly reasons about geometry based on static scene assumption, which can be easily affected by dynamic objects and occlusion. DynamicDepth~\cite{dynamicdepth} attempts to tackle dynamic scenes, but relies on depth priors and segmentation masks of moving objects when reasoning. To sidestep the issue of dynamic objects in multi-frame inference encountered by these methods, we return to the traditional paradigm~\cite{Patil,tcdepth}, which aggregates multi-frame features directly. However, direct feature fusion without alignment harms the learning of temporal representations, since feature embeddings at the same location in neighbor frames are usually irrelevant, making it difficult for models to leverage the correlation of multiple frames. Hence, we regard feature alignment as a reasonable and necessary step before aggregation. As revealed in several video enhancement tasks like video super-resolution~\cite{BasicVSR,BasicVSRp} and frame interpolation~\cite{rife,ifrnet,amt,atca}, inter-frame feature alignment can be implemented by warping through optical flow. This inspires us to employ a flow-based VFI model to concurrently conduct temporal augmentation and provide explicit correspondence for feature alignment. Moreover, the VFI model can reason out occlusion mask between neighbor frames, helping to alleviate inter-frame occlusion problem in feature fusion. Based on above analysis, we design a VFI-assisted multi-frame fusion module for multi-frame depth estimation, as depicted in \cref{fig:fuse_module}. 

Finally, we build a unified framework to bilaterally connect single- and multi-frame depth, as demonstrated in \cref{fig:overview}(b). The direct way is by depth consistency, usually for geometric regularization or distillation in self-supervised MDE. ManyDepth conducts unilateral distillation from single-frame model to multi-frame one. DynamicDepth extends it to a bilateral mutual reinforcement, but only in the regions of moving objects. Differently, our framework applies the bilateral consistency on the whole image. For an original target image, we use a standard view depth consistency (SVDC) loss between its two depth maps predicted from the single- and multi-frame models. We also include classic data augmentation strategies, categorized as spatial augmentation. Specifically, we combine resizing-cropping with image rotation, forming a generic affine transformation. Then, we give the formulation of rectified camera pose for affine-augmented view synthesis using the cue of relative object size~\cite{planedepth,cue}. For the depth map from augmented target view, we enforce two scale-aware depth consistency (SADC) losses between it and above two standard view depths, respectively. SADC stems from the relative size cue, which assumes that when an image is zoomed up by a factor, the depth decreases by the same factor. It guides depth model to learn consistent geometric scale relationship from scene variations. The SVDC loss and two SADC losses are termed as a triplet depth consistency loss. 

Our framework is named \textbf{Mono-ViFI}, since the self-supervised \textbf{Mono}cular depth is largely boosted by \textbf{Vi}deo \textbf{F}rame \textbf{I}nterpolation. To our best knowledge, we are the first to leverage VFI for the improvement of self-supervised MDE. The main contributions of this paper are summarized as follows:
\begin{itemize}
  \item We propose Mono-ViFI, a unified learning framework to bilaterally connect self-supervised single- and multi-frame MDE. In Mono-ViFI, the single- and multi-frame models can share weights, making the unified framework more compact and memory-efficient.
  \item We conduct temporal augmentation through flow-based VFI, and develop a novel VFI-assisted multi-frame fusion module, which utilizes motion and occlusion from flow-based VFI model to improve multi-frame depth.
  \item We incorporate image affine transformation as spatial augmentation, along with a triplet depth consistency loss for regularization and distillation.
  \item Experiments show that Mono-ViFI can achieve state-of-the-art performance. Moreover, our approach is orthogonal to architecture design, allowing current advanced depth networks to be further improved by seamlessly integrating them with Mono-ViFI.
\end{itemize}

\section{Related Work}
% \label{sec:formatting}

\subsection{Self-supervised Monocular Depth Estimation}
Self-supervised learning paradigm for monocular depth estimation can be categorized into stereo training and monocular training according to the data source.

Stereo training leverages the left-right consistency in stereo image pairs. Garg \etal~\cite{garg} first consider the training as a view synthesis problem with a photometric consistency loss between stereo pairs. Building upon this idea, Godard \etal~\cite{godard} introduce a left-right disparity consistency loss. Then, it is further extended by using depth hints~\cite{hint}, data augmentation~\cite{epcdepth,planedepth}, knowledge distillation~\cite{pilzer,epcdepth,planedepth} and plane-based scene representation~\cite{falnet, planedepth}.

Monocular training leverages the temporal consistency in monocular video frames. Zhou \etal~\cite{sfmlearner} propose a pioneering method which jointly learns depth and camera ego-motion from monocular videos in a self-supervised way. Then, Bian \etal~\cite{scdepth} propose a scale-consistent depth and ego-motion estimation approach by adding a depth consistency loss, which reduces scale drift of estimated depth but decreases the accuracy. Later, Godard \etal~\cite{mono2} propose Monodepth2, a strong baseline for most follow-up works, which employs a minimum reprojection loss to mitigate occlusion issues and an auto-masking loss to filter out moving objects with similar velocity as the camera. Furthermore, some works introduce multi-task learning to combine depth estimation with optical flow~\cite{geonet,glnet,cc,lee}, semantic segmentation~\cite{sdo,fgs,depthsegnet} and instance segmentation~\cite{iapc}. Others~\cite{packnet,hrdepth,diffnet,monovit,das,brnet,lite-mono,daccn,bdedepth} try to improve the network architectures for depth estimation. However, these methods only utilize the geometric constrains between existing fixed camera views in video datasets. This insufficient guidance may block current depth models from unleashing their full capabilities. To address it, we conduct temporal augmentation through VFI in this paper.

\subsection{Multi-frame Monocular Depth}
Above single-frame depth models do not leverage any temporal information for inference, which limits the performance to some degree. Regarding multi-frame depth prediction, early works adopt test-time refinement~\cite{glnet,cvd,comoda}, use spatial-temporal attention~\cite{tcdepth}, or combine traditional depth networks with recurrent layers~\cite{wang,zhang} to process sequence information. Another broad group of methods~\cite{manydepth,dynamicdepth,depthformer,movedepth} utilize cost volume from stereo matching, which explicitly encode and reason from geometric constraints by combining a multi-frame cost volume with single-frame features. On the one hand, this type of methods based on ManyDepth~\cite{manydepth} usually meet with the problem of dynamic scenes. To sidestep this issue, we develop a VFI-assisted multi-frame fusion depth model to align and aggregate multi-frame features by optical flow and occlusion masks. On the other hand, ManyDepth uses single-frame depth prior to boost multi-frame depth estimation. However, this improvement is unidirectional. DynamicDepth~\cite{dynamicdepth} extends it to a mutual reinforcement, but only targets on dynamic objects. Differently, we propose a unified framework to bilaterally connect single- and multi-frame depth in a more compact and memory-efficient manner.

\section{Problem Setup}
\setlength{\abovedisplayskip}{2mm}
\setlength{\belowdisplayskip}{2mm}
\label{sec:setup} 
Self-supervised MDE in the monocular training case leverages the geometric constraints from video sequences. Specifically, for a training triplet $\{I_{t-2}, I_t, I_{t+2}\}$ which contains three consecutive frames, we use $I_t$ as the target view and the other two as the source views $\{I_s\}_{s\in \{t-2,~t+2\}}$ to implement view synthesis. Here the temporal notation $t\pm2$ instead of $t\pm1$ is for the convenience of following description. Also, a pose network is utilized to estimate the camera pose from the target view to the source view, denoted as $\boldsymbol{T}_{t \rightarrow s}=[\boldsymbol{R}_{t \rightarrow s}|\boldsymbol{t}_{t \rightarrow s}]$, where $\boldsymbol{R}_{t \rightarrow s}$ and $\boldsymbol{t}_{t \rightarrow s}$ represent the rotation and translation components. Formally, the target depth $D_t$ and the camera pose $\boldsymbol{T}_{t \rightarrow s}$ can be derived from:
\begin{equation}
D_t=\mathcal{D}(\mathcal{E}(I_t)), ~~~ \boldsymbol{T}_{t \rightarrow s}=\mathcal{P}(I_t,I_s),
\end{equation}
where $\mathcal{E}$ and $\mathcal{D}$ represent the encoder and decoder, respectively, of a single-frame depth network, while $\mathcal{P}$ denotes the pose network. Then, the reconstructed view of $I_t$ from the source view $I_s$ can be obtained by:
\begin{equation}
I_{s \rightarrow t}=I_s\big\langle proj(D_t,\boldsymbol{T}_{t \rightarrow s},\boldsymbol{K}) \big\rangle,
\end{equation}
where $I_{s \rightarrow t}$ is the reconstructed target image, $\langle \rangle$ means the differentiable bilinear sampling operator, $proj()$ returns the 2D coordinates of the pixels in $I_t$ after they are reprojected to the viewpoint of $I_s$, and $\boldsymbol{K}$ is the camera intrinsics matrix assumed to be known. Finally, the per-pixel photometric loss between $I_{s \rightarrow t}$ and $I_t$ is used to optimize the depth and pose networks in an end-to-end manner. We follow Monodepth2~\cite{mono2} to adopt the combination of L1 distance and SSIM~\cite{ssim} as the photometric error function and take the minimum across all the source views at each pixel to handle occlusion, which is formulated as:
\begin{equation}
\label{eq3}
L_{pe}(t)=\min\limits_{s}~ \frac{\alpha}{2}(1-{\rm{SSIM}}(I_t,I_{s \rightarrow t})) + (1-\alpha)||I_t,I_{s \rightarrow t}||_1,
\end{equation}
where $\alpha=0.85$. Besides, the edge-aware smoothness loss is employed to cope with depth discontinuities:
\begin{equation} 
\label{eq5}
L_{sm}(t)=|\partial_x d_t^*|e^{-|\partial_x I_t|}+|\partial_y d_t^*|e^{-|\partial_y I_t|},
\end{equation}
where $d_t^*= d_t/\overline{d_t}$ is the mean-normalized inverse depth to discourage shrinking~\cite{direct}. In summary, the standard self-supervised loss on $I_t$ is:
% \begin{small}
\begin{equation} 
\label{eq6}
L_{ss}(t)=\mathcal{L}(D_t,\boldsymbol{T}_{t \rightarrow s},I_t,I_s)=\mu L_{pe}(t) + \gamma L_{sm}(t),
\end{equation}
% \end{small}
where $\mu$ is the auto-masking in~\cite{mono2} and $\gamma=0.001$.

\section{Methodology}

In this paper, we propose novel Mono-ViFI for self-supervised monocular training, which bilaterally connects single- and multi-frame depth models into a unified learning framework. With flow-based VFI model, we introduce temporal augmentation and develop VFI-assisted multi-frame fusion depth, which are presented in \cref{sec:4.1} and \cref{sec:4.2}, respectively. Besides, our framework incorporates spatial data augmentation and a triplet depth consistency loss, which are separately described in \cref{sec:4.3} and \cref{sec:4.4}. Finally, we give the overall training objective in \cref{sec:4.5}. The overview of our Mono-ViFI is depicted in \cref{fig:overview}.

\subsection{Temporal Augmentation with VFI}
\label{sec:4.1}
\setlength{\abovedisplayskip}{2mm}
\setlength{\belowdisplayskip}{2mm}
Firstly, we generate more new frames in the temporal dimension based on existing video frames for temporal augmentation. To achieve this, we adopt a pretrained flow-based VFI model, such as IFRNet~\cite{ifrnet}, which takes two temporally adjacent frames $\{I_0, I_2\}$ as input to synthesize an intermediate frame $I_1$. The frame interpolation process can be formulated as:
\begin{gather} 
F_{1 \rightarrow 0},~F_{1 \rightarrow 2}, ~M_{1} = \mathcal{H}(I_0, I_2), \\
I_{0 \rightarrow 1}=\omega(I_0, F_{1 \rightarrow 0}),~~~~I_{2 \rightarrow 1}=\omega(I_2, F_{1 \rightarrow 2}), \\
I_1 = M_1 \odot I_{0 \rightarrow 1} + (1-M_1)\odot I_{2 \rightarrow 1}, \label{eq:vfi2}
\end{gather}
where $\mathcal{H}$ is the VFI model, $F_{1 \rightarrow 0}$ and $F_{1 \rightarrow 2}$ are the intermediate optical flow, $\omega$ represents backward warping. $I_{0 \rightarrow 1}$ and $I_{2 \rightarrow 1}$ denote the warped images to the intermediate position, while $M_1$ ranging from 0 to 1 is the merge mask adjusting the mixing ratio according to bidirectional occlusion information. Here, $\odot$ indicates the element-wise multiplication. Considering a triplet $\{I_{t-2}, I_t, I_{t+2}\}$ with three consecutive frames as described in \cref{sec:setup}, we use the VFI model to synthesize $I_{t-1}$ from $\{I_{t-2}, I_t\}$ and $I_{t+1}$ from $\{I_t, I_{t+2}\}$, as shown in \cref{fig:overview}(a). Then, we regard $I_{t-1}$ and $I_{t+1}$ as another two target frames. All three target frames have common sources of $\{I_{t-2}, I_{t+2}\}$. The following operations and losses are the same for each of them, and we only take $I_{t}$ for instance.

\subsection{VFI-assisted Multi-frame Fusion Depth} 
\label{sec:4.2}
\begin{figure*}[t]
\centering
%   \fbox{\rule{0pt}{2in} \rule{0.9\linewidth}{0pt}}
\includegraphics[width=1.0\linewidth]{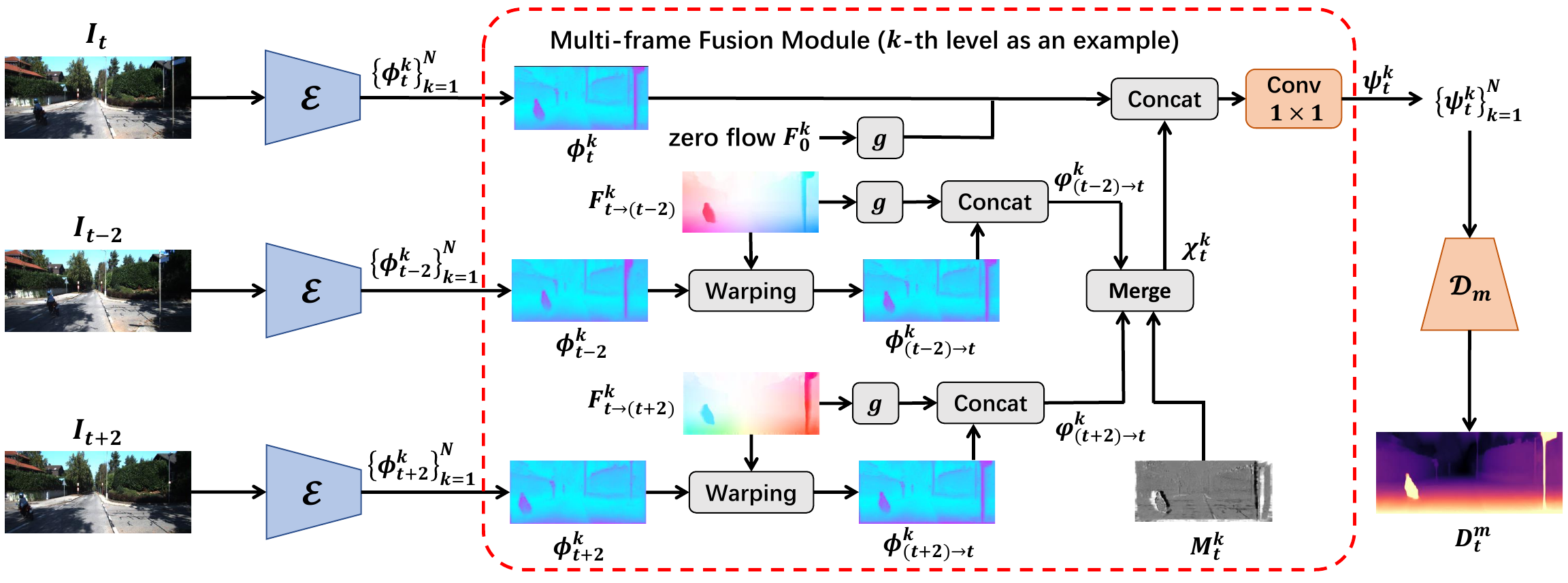}
\caption{Our VFI-assisted multi-frame fusion depth model. The intermediate optical flow and occlusion merge mask provided by the VFI network can also be used to align and aggregate multi-frame features in an explicit manner.}
\label{fig:fuse_module}
\vspace{-6mm}
\end{figure*}

\setlength{\abovedisplayskip}{2mm}
\setlength{\belowdisplayskip}{2mm}
Apart from temporal augmentation, our employed VFI model can also provide other inter-frame information, such as optical flow and occlusion merge masks. This enables us to design a novel VFI-assisted multi-frame fusion module for multi-frame depth prediction, as demonstrated in \cref{fig:fuse_module}. The fusion module consists of two parts, \ie, motion-aware feature alignment (MAFA) and occlusion-alleviated feature fusion (OAFF). Taking original triplet $\{I_{t-2}, I_t, I_{t+2}\}$ for example, we implement forward process of the VFI model to interpolate the middle position $t$ of $\{I_{t-2}, I_{t+2}\}$ in \cref{eq9}, yielding intermediate flow $F_{t \rightarrow (t-2)}, F_{t \rightarrow (t+2)}$ and merge mask $M_t$ to align and aggregate multi-frame features.
\begin{equation} \label{eq9}
    F_{t \rightarrow (t-2)}, F_{t \rightarrow (t+2)}, M_t = \mathcal{H}(I_{t-2}, I_{t+2}).
\end{equation}
% which can be used to align and aggregate multi-frame features. 

\noindent \textbf{Multi-frame Feature Extraction.} Our fusion module accepts multi-frame features extracted by the same encoder, which is shared with the single-frame depth encoder $\mathcal{E}$. The extraction process can be written as:
\begin{equation}
    \{\phi_{t-2}^k, \phi_{t}^k, \phi_{t+2}^k\}_{k=1}^{N} = \{\mathcal{E}(I_{t-2}), \mathcal{E}(I_{t}), \mathcal{E}(I_{t+2})\},
\end{equation}
where $\{\phi_{t-2}^k, \phi_{t}^k, \phi_{t+2}^k\}_{k=1}^{N}$ are the multi-level features of the three frames and $N$ is the number of feature levels.

\noindent \textbf{Motion-aware Feature Alignment (MAFA).} For the $k$-th level, we use bidirectional optical flow to align the neighbor features $\phi_{t-2}^k$ and $\phi_{t+2}^k$ to the target position $t$, generating $\phi_{(t-2) \rightarrow t}^k$ and $\phi_{(t+2) \rightarrow t}^k$, respectively, formulated as:
\begin{equation}
\begin{aligned}
    \phi_{(t-2) \rightarrow t}^k&=\omega(\phi_{t-2}^k,~ F^k_{t \rightarrow (t-2)}), \\
    \phi_{(t+2) \rightarrow t}^k&=\omega(\phi_{t+2}^k,~ F^k_{t \rightarrow (t+2)}),
\end{aligned}  
\end{equation}
where $F^k_{t \rightarrow (t-2)}$ and $F^k_{t \rightarrow (t+2)}$ denote the corresponding flows of the $k$-th level, downsampled from $F_{t \rightarrow (t-2)}$ and $F_{t \rightarrow (t+2)}$, respectively. Inspired by recent 2D implicit neural representation~\cite{LIIF}, we also concatenate the aligned features with encoded optical flow, which can be denoted as:
\begin{equation}
\begin{aligned}
    \varphi_{(t-2) \rightarrow t}^k&={\rm Concat}(\phi_{(t-2) \rightarrow t}^k, ~g(F^k_{t \rightarrow (t-2)})), \\
    \varphi_{(t+2) \rightarrow t}^k&={\rm Concat}(\phi_{(t+2) \rightarrow t}^k, ~g(F^k_{t \rightarrow (t+2)})), \\
    \varphi_{t}^k&={\rm Concat}({\phi}_{t}^k, ~g(F^k_{0})),
\end{aligned}  
\end{equation}
where $\rm Concat$ means feature concatenation, $F^k_{0}$ denotes zero flow and $g$ is the fourier positional encoding function~\cite{ffl,nerf} (see in supplementary material) to transform optical flow to high-frequency signal. Actually, optical flow indicates the relative coordinate of a feature vector in the warped feature map with respect to the original one. When incorporating it, the multi-frame model can adaptively adjust the feature learning through motion-aware alignment.
 
\noindent \textbf{Occlusion-alleviated Feature Fusion (OAFF).} Due to the ubiquitous inter-frame occlusion, aggregating the aligned features directly through summation or concatenation can deteriorate the performance. Therefore, we use the $k$-th level merge mask $M_t^k$ downsampled from $M_t$ to mitigate this issue, just like the image merging process of VFI in \cref{eq:vfi2}:
\begin{equation}
    \chi_t^k = M_t^k \odot \varphi_{(t-2) \rightarrow t}^k + (1-M_t^k) \odot \varphi_{(t+2) \rightarrow t}^k,
\end{equation}
where $\chi_t^k$ is the occlusion-alleviated feature merged from the temporal positions $t-2$ and $t+2$. It is then concatenated with the feature $\varphi_{t}^k$ in the position $t$. And we can obtain the final feature after fusion by:
\begin{equation}
    \psi_{t}^k = {\rm Conv_{1\times1}}({\rm Concat}(\varphi_{t}^k,~ \chi_t^k)),
\end{equation}
where $\rm Conv_{1\times1}$ is a $1\times1$ convolutional layer to transform the feature into the same channel as the original feature $\phi_t^k$, enabling the multi-frame depth decoder to employ the same architecture as the single-frame one.

\noindent \textbf{Multi-frame Depth Decoder.} To translate the multi-level fusion features into a depth map, a decoder is required. Concretely, we use the same architecture as the single-frame depth decoder, but the parameters are not shared. This encourages the single- and multi-frame models to decode the features in their individual ways without conflict. Finally, the fusion features $\{\psi_{t}^k\}_{k=1}^N$ are processed by a multi-frame depth decoder $\mathcal{D}_m$ to generate the fusion depth $D_{t}^{m}$ of the target frame $I_t$, which can be written as: 
\begin{equation}
    D_{t}^m =\mathcal{D}_m(\{\psi_{t}^k\}_{k=1}^N).
\end{equation}
Note that the multi-frame depth $D_{t}^m$ also corresponds to a self-supervised loss like \cref{eq6}, which is denoted as:
\begin{equation} \label{eq20}
L_{ss}^m(t)=\mathcal{L}(D_{t}^m,\boldsymbol{T}_{t \rightarrow s},I_t,I_s).
\end{equation}
% Recall that we have respectively obtained $D_t$ and $\widetilde{D}_t$ from the original image $I_t$ and the augmented image $\widetilde{I}_t$ through the single-frame depth model. And $\widetilde{D}_t$ is inversely transformed to $\widehat{D}_t=\mathcal{A}^{-1}(\widetilde{D}_t)$. Thus, similarly to Eq.~\ref{eq9}, we can derive a spatial augmentation depth consistency loss $L_{ac}^m(t)$ from $\{D_{t}^m, \widetilde{D}_t\}$ and a standard view depth consistency loss $L_{vc}(t)$ from $\{D_{t}^m, D_{t}\}$, which can be written as:
% \begin{equation}  \label{eq21}
%     L_{ac}^m(t)={\rm SI}(M_{ac} \odot D_{t}^m, M_{ac} \odot f_s \widehat{D}_t),
% \end{equation}
% \begin{equation} \label{eq22}
%     L_{vc}(t)={\rm SI}(D_{t}^m, D_{t}).
% \end{equation}
% The two losses are depicted in \Cref{fig:overview}(b), which can correct the multi-frame model in regions with inaccurate motion and occlusion by leveraging information from spatial augmentation. In turn, the powerful multi-frame model can also guide the learning of single-frame model. 

% For the synthesized target frames $I_{t-1}$ and $I_{t+1}$, we can derive their multi-frame fusion depths $D_{t-1}^m$ and $D_{t+1}^m$ from $\{I_{t-2}, I_{t-1}, I_{t}\}$ and $\{I_{t}, I_{t+1}, I_{t+2}\}$ in the same way. And their corresponding losses are similar to Eqs.~\ref{eq20}-\ref{eq22}.

\subsection{Spatial Augmentation}
\label{sec:4.3}
\setlength{\abovedisplayskip}{2mm}
\setlength{\belowdisplayskip}{2mm}
Orthogonal to temporal augmentation, we also include classic data augmentation to spatially expand the data diversity, which is categorized as spatial augmentation in this paper. Resizing-cropping is a commonly used and effective augmentation approach. Based on this operation, we introduce image rotation, extending it to a generic affine transformation (cropping is actually image translation). Specifically, we follow~\cite{planedepth} to consider that the augmented image and the original image are captured by the same camera system whose intrinsics are fixed. For resizing operation, it is assumed that when an image is zoomed up by a factor $f_s \ge 1$, the relative size of an object increases by the same factor $f_s$ and its depth decreases by $f_s$ simultaneously. This is inspired from a vital monocular cue~\cite{cue} that the closer an object is, the larger its relative size is. Our affine transformation $\mathcal{A}$ is applied to both the target image $I_t$ and the source image $I_s$, generating $\widetilde{I}_t=\mathcal{A}(I_t)$ and $\widetilde{I}_s=\mathcal{A}(I_s)$, respectively. For the augmented target view, we can obtain its depth through the single-frame model by $\widetilde{D}_t = \mathcal{D}(\mathcal{E}(\widetilde{I}_t))$. To conduct view synthesis on augmented images, the original camera pose $\boldsymbol{T}_{t \rightarrow s}=[\boldsymbol{R}_{t \rightarrow s}|\boldsymbol{t}_{t \rightarrow s}]$ from target to  source should be rectified to $\boldsymbol{\widetilde{T}}_{t \rightarrow s}=[\boldsymbol{\widetilde{R}}_{t \rightarrow s}|\boldsymbol{\widetilde{t}}_{t \rightarrow s}]$ as done in~\cite{planedepth}:
\begin{equation}
\boldsymbol{\widetilde{R}}_{t \rightarrow s}=\boldsymbol{R}_c\boldsymbol{R}_{t \rightarrow s}\boldsymbol{R}_c^{-1},~~~~\boldsymbol{\widetilde{t}}_{t \rightarrow s}=\boldsymbol{R}_c\boldsymbol{t}_{t \rightarrow s},
\label{eq7}
\end{equation}
where $\boldsymbol{R}_c$ is the rectification matrix determined only by the camera intrinsics and the augmentation parameters. Different from \cite{planedepth} calculating $\boldsymbol{R}_c$ under resizing-cropping, we give its formulation and derivation under a more general affine transformation, which can be seen in our supplementary. Therefore, we can derive the self-supervised loss $\widetilde{L}_{ss}(t)$ on the augmented images similarly to \cref{eq6}:
\begin{equation}  \label{eq8}
\widetilde{L}_{ss}(t)=\mathcal{L}(\widetilde{D}_{t},\boldsymbol{\widetilde{T}}_{t \rightarrow s}, \widetilde{I}_t,\widetilde{I}_s).
\end{equation}
Note that there is a mask to exclude the possible invalid pixels in $\widetilde{I}_{t}$ when calculating $\widetilde{L}_{ss}(t)$ due to image rotation.

\subsection{Triplet Depth Consistency}
\label{sec:4.4}
\setlength{\abovedisplayskip}{2mm}
\setlength{\belowdisplayskip}{2mm}
Now we have single-frame depths $D_t$ and $\widetilde{D}_t$ of the original image $I_t$ and the augmented image $\widetilde{I}_t$, along with multi-frame depth $D_t^m$ of $I_t$. For better knowledge distillation and regularization, our unified learning framework applies a triplet depth consistency loss among the three depth maps, including a standard view depth consistency (SVDC) loss and two scale-aware depth consistency (SADC) losses, which are shown in \cref{fig:overview}(b). The SVDC loss $L_{sv}(t)$ conducts mutual distillation between single- and multi-frame models, promoting their mutual reinforcement, which can be denoted as:
\begin{equation} \label{eq19}
    L_{sv}(t)={\rm SI}(D_{t}^m, D_{t}),
\end{equation}
where $\rm SI()$ is the widely used scale-invariant error function proposed in~\cite{eigen}. Besides, we use the depth $\widetilde{D}_t$ under spatial augmentation to enforce two SADC losses, \ie, $L_{sa}(t)$ between $\widetilde{D}_t$ and $D_t$, and $L_{sa}^m(t)$ between $\widetilde{D}_t$ and $D_t^m$. To this end, we first restore $\widetilde{D}_t$ to the original image view, generating $\widehat{D}_t=\mathcal{A}^{-1}(\widetilde{D}_t)$ in \cref{fig:overview}(b), where $\mathcal{A}^{-1}$ is the inverse transformation of affine augmentation $\mathcal{A}$. To exclude the invalid pixels in $\widehat{D}_t$, we adopt a mask $M_{sa}$ and calculate $L_{sa}(t)$ and $L_{sa}^m(t)$ through:
\begin{gather} 
    L_{sa}(t)={\rm SI}(M_{sa} \odot D_t, M_{sa} \odot f_s \widehat{D}_t), \\
    L_{sa}^m(t)={\rm SI}(M_{sa} \odot D_{t}^m, M_{sa} \odot f_s \widehat{D}_t),
\end{gather}
where the factor $f_s$ is due to the assumption in \cref{sec:4.3} that the depth of an object in $I_t$ is $f_s$ times of that in $\widetilde{I}_t$. Our SADC constrains the 3D geometric relationship between the scenes of $I_t$ and $\widetilde{I}_t$, providing scale-aware regularization. Combining SVDC and SADC, we have the triplet depth consistency loss as:
\begin{equation}  \label{eq22}
    L_{tc}(t)=L_{sv}(t)+L_{sa}(t)+L_{sa}^m(t).
\end{equation}

\subsection{Training Objective}
\label{sec:4.5}
\setlength{\abovedisplayskip}{2mm}
\setlength{\belowdisplayskip}{2mm}
For target position $t$, we have already obtained the self-supervised losses $L_{ss}(t)$, $L_{ss}^m(t)$ and $\widetilde{L}_{ss}(t)$ in \cref{eq6,eq20,eq8}, respectively. Besides, our triplet depth consistency loss $L_{tc}(t)$ is acquired in \cref{eq22}. Therefore, the total loss on position $t$ can be summed as:
\begin{equation}
    L(t)=L_{ss}(t)+L_{ss}^m(t)+\widetilde{L}_{ss}(t)+\lambda L_{tc}(t),
\end{equation}
where $\lambda$ is a hyper-parameter set to $0.2$ in this paper. For $t-1$ and $t+1$, we can derive their losses $L(t-1)$ and $L(t+1)$ in the same way. In summary, the overall loss function of our Mono-ViFI is $L=L(t-1)+L(t)+L(t+1)$.

\begin{table*}[t]
    \begin{center}
    \caption{Performance comparison on KITTI~\cite{kitti} benchmark using the raw ground truth, with a resolution of $640 \times 192$. The best results in each subsection are in \textbf{bold}; second best are \underline{underlined}. For MACs of multi-frame models, we use the formulation A/B. A denotes the MACs when predicting for a single frame; B means the average MACs per frame when processing a long video sequence. \#Params and MACs of our multi-frame models include the VFI network.}
    \label{tab1}
    \vspace{-5mm}
    \resizebox{1.0\textwidth}{!}{
    \begin{tabular}{|c|c|c|c||c|c|c|c|c|c|c|}
    \hline
    Method  & Test frames &\#Params & MACs & \cellcolor{col1}Abs Rel & \cellcolor{col1}Sq Rel & \cellcolor{col1}RMSE  & \cellcolor{col1}RMSE log & \cellcolor{col2}$\delta_1 $ & \cellcolor{col2}$\delta_2$ & \cellcolor{col2}$\delta_3$\\
    \hline
    \hline
    PackNet-SfM~\cite{packnet} & 1 & 128M & 205G & 0.111 & 0.785 & 4.601 & 0.189 & 0.878 & 0.960 &  0.982 \\
    HR-Depth~\cite{hrdepth} & 1 & 16.9M  & 31.6G & 0.109 & 0.792 & 4.632 & 0.185 & 0.884 & 0.962 & 0.983 \\
    R-MSFM6~\cite{r-msfm} & 1 & \underline{3.8M} & 31.2G & 0.112 & 0.806 & 4.704 & 0.191 & 0.878 & 0.960 & 0.981 \\
    BRNet~\cite{brnet} & 1 & 19.1M & 33.7G & 0.105 & 0.698 & 4.462 & 0.179 & 0.890 & 0.965 & \underline{0.984} \\
    MonoViT~\cite{monovit} & 1 & 27M & - & \underline{0.099} & 0.708 & 4.372 & 0.175 & \underline{0.900} & 0.967 & \underline{0.984} \\
    DIFFNet~\cite{diffnet} & 1 & 12.2M & 23.4G & 0.102 & 0.764 & 4.483 & 0.180 & 0.896 & 0.965 & 0.983 \\
    SD-SSMDE~\cite{sd-ssmde} & 1 & - & - & 0.100 & 0.661 & 4.264 & 0.172 & 0.896 & 0.967 & \textbf{0.985} \\
    MUSTNet~\cite{mustnet} & 1 & - & - & 0.106 & 0.763 & 4.562 & 0.182 & 0.888 & 0.963 & 0.983 \\
    DepthSegNet~\cite{depthsegnet} & 1 & 30.3M & - & 0.104 & 0.690 & 4.473 & 0.179 & 0.886 & 0.965 & \underline{0.984}  \\
    DualRefine~\cite{dualrefine} & 1 & - & - & 0.103 & 0.776 & 4.491 & 0.181 & 0.894 & 0.965 & 0.983 \\
    DaCCN~\cite{daccn} & 1 & 13M & - & \underline{0.099} & 0.661 & 4.316 & 0.173 & 0.897 & 0.967 & \textbf{0.985} \\
    \hdashline  
    Monodepth2 (ResNet18)~\cite{mono2}& 1 & 14.3M & \underline{8G} & 0.115 & 0.903 & 4.863 & 0.193 & 0.877 & 0.959 & 0.981\\
    \textbf{Mono-ViFI (ResNet18)}& 1 & 14.3M & \underline{8G} &0.105 & 0.708 & 4.446 & 0.179 & 0.887 & 0.965 & \underline{0.984} \\   
    \hdashline 
    Lite-Mono~\cite{lite-mono}& 1  & \textbf{3.1M} & \textbf{5.1G} & 0.107 & 0.765 & 4.561 & 0.183 & 0.886 & 0.963 & 0.983 \\
    \textbf{Mono-ViFI (Lite-Mono)}& 1  & \textbf{3.1M} & \textbf{5.1G} & 0.103 & 0.680 & 4.349 & 0.176 & 0.891 & 0.966 & \textbf{0.985} \\    
    \hdashline 
    D-HRNet~\cite{radepth} & 1 & 10M & 10.8G& 0.102  & 0.760 & 4.479 & 0.179 & 0.897  & 0.965 & 0.983\\
    RA-Depth (D-HRNet)~\cite{radepth}& 1  & 10M &10.8G & \textbf{0.096} & \underline{0.632} & \underline{4.216} & \underline{0.171} & \textbf{0.903} & \underline{0.968} & \textbf{0.985} \\
    \textbf{Mono-ViFI (D-HRNet)} & 1 & 10M & 10.8G& \textbf{0.096}  & \textbf{0.627} & \textbf{4.179} & \textbf{0.170} & \textbf{0.903}  & \textbf{0.969} & \textbf{0.985}\\
    \hline 
    Patil \etal~\cite{Patil} & N & - & -/- & 0.111 & 0.821 & 4.650 & 0.187 & 0.883 & 0.961 & 0.982\\
    ManyDepth (ResNet18)~\cite{manydepth} & 2 (-1, 0)& 26.9M &\textbf{15.1G}/\underline{13.7G} & 0.098 & 0.770 & 4.459 & 0.176 & 0.900 & 0.965 & 0.983\\
    DynamicDepth~\cite{dynamicdepth} & 2 (-1, 0)& - & -/- & 0.096 & 0.720 & 4.458 & 0.175 & 0.897 & 0.964 & 0.984 \\ 
    DepthFormer~\cite{depthformer} & 2 (-1, 0)& 28.7M & 175G/-~~~~~ & \underline{0.090} & 0.661 & \underline{4.149} & 0.175 & 0.905 & \underline{0.967} & 0.984 \\  
    MOVEDepth~\cite{movedepth} & 2 (-1, 0)& 28.2M &\underline{20.2G}/19.2G & 0.094 & 0.704 & 4.389 & 0.175 & 0.902 & 0.965 & 0.983 \\
    DualRefine~\cite{dualrefine} & 2 (-1, 0) & - & - & \textbf{0.087} & 0.698 & 4.234 & \underline{0.170} & \textbf{0.914} & 0.967 & 0.983 \\
    TC-Depth~\cite{tcdepth} & 3 (-1, 0, 1) & - & -/- & 0.103 & 0.746 & 4.483 & 0.185 & 0.894 & - & 0.983\\
    \textbf{Mono-ViFI (ResNet18)} & 3 (-1, 0, 1) & 17.9M &25.8G/16.9G & 0.099 & 0.661 & 4.321 & 0.174 & 0.898 & 0.966 & \underline{0.985}\\
    \textbf{Mono-ViFI (Lite-Mono)} & 3 (-1, 0, 1)& \textbf{5.9M} & 21.9G/\textbf{13.3G} & 0.099 & \underline{0.648} & 4.256 & 0.172 & 0.898 & \underline{0.967} & \underline{0.985}\\
    \textbf{Mono-ViFI (D-HRNet)} & 3 (-1, 0, 1)& \underline{12.9M} & 35.7G/19.4G & 0.091 & \textbf{0.589} & \textbf{4.088} & \textbf{0.166} & \underline{0.912} & \textbf{0.969} & \textbf{0.986}\\
    \hline
    \end{tabular}}
    \vspace{-4mm}
    \end{center}
    \end{table*}

\section{Experiments}

\begin{figure*}[t]
    \centering
    %   \fbox{\rule{0pt}{2in} \rule{0.9\linewidth}{0pt}}
    \includegraphics[width=1.0\linewidth]{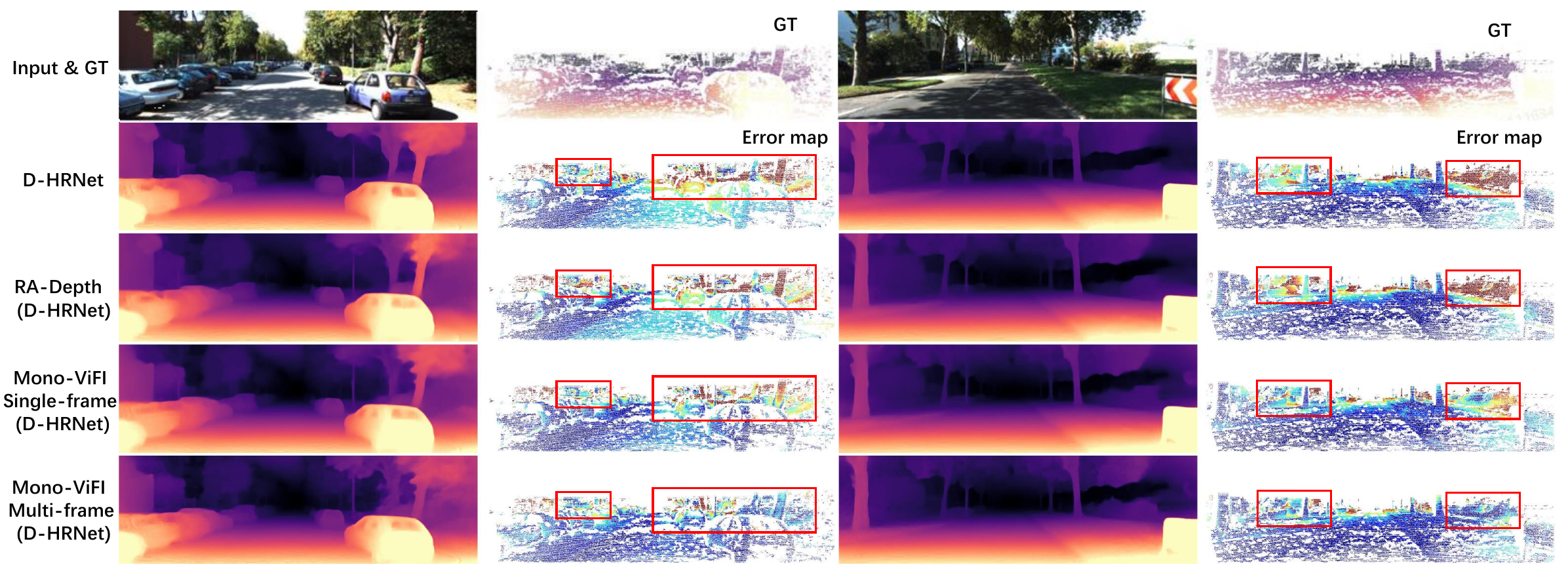}
    \vspace{-5mm}
    \caption{Qualitative results on KITTI. Error maps in columns 2 and 4 show the Abs Rel error compared to the improved ground truth~\cite{im-gt} from good (blue) to bad (red).}
    \label{fig:vis}
    \vspace{-6mm}
    \end{figure*}

\subsection{Implementation Details}
\noindent \textbf{Datasets.} We train and evaluate the models on two widely used outdoor urban datasets of KITTI~\cite{kitti} and Cityscapes~\cite{city}. The depth labels of KITTI are collected with LiDARs. For fair comparison, we use the Eigen split~\cite{eigen2} and follow the pre-processing operation in~\cite{mono2}, generating 39,810 monocular triplets for training and 697 frames for evaluation. To reduce the influence of noise in the sparse depth from Velodyne, we also evaluate using $93\%$ of the Eigen split with the improved ground truth from~\cite{im-gt}, which contains 652 test frames. In regard to Cityscapes, following~\cite{sfmlearner,manydepth}, we preprocess the monocular sequences into triples, producing 69,731 training images. And we evaluate on 1,525 test images using the ground truth provided by \cite{manydepth}. For generalization comparison, KITTI models are evaluated on Cityscapes and Make3D~\cite{make3d}. Make3D is a small outdoor dataset containing 134 test images with aligned depth information. 

\noindent \textbf{Flow-based VFI Model.} We choose IFRNet~\cite{ifrnet} as our VFI model due to its efficiency. In practice, we use the large model of IFRNet for training and replace it with the small model when evaluating for multi-frame fusion depth. Please see ablation about their gap in supplementary. Both large and small IFRNets are pretrained on KITTI/Cityscapes and frozen in Mono-ViFI training. The pretraining data split is the same as self-supervised depth training. Each original triplet in depth training is directly used for VFI training. Besides, we employ another two flow-based VFI networks, RIFE~\cite{rife} and AMT~\cite{amt}, to verify the robustness of our method to different VFI models.

\noindent \textbf{Depth and Pose Networks.} We experiment with three depth architectures, including ResNet18~\cite{resnet} based network in Monodepth2~\cite{mono2}, Lite-Mono~\cite{lite-mono}, and D-HRNet proposed in RA-Depth~\cite{radepth}. For the pose network, we follow~\cite{mono2} to use a ResNet18 based model which takes a pair of images as input and predict a 6-DoF relative pose. All their Encoders are pretrained on ImageNet~\cite{imagenet}. 

\noindent \textbf{Evaluation Metrics.} For depth evaluation, standard metrics from~\cite{eigen,eigen2} are adopted, including error-based metrics for which lower is better (Abs Rel, Sq Rel, RMSE, RMSE log) and accuracy-based metrics for which higher is better ($\delta_1< 1.25$, $\delta_2< 1.25^2$, $\delta_3< 1.25^3$). 

\subsection{Performance Comparison}
\noindent \textbf{KITTI Results.} Quantitative comparison results on KITTI under $640\times192$ resolution are listed in \Cref{tab1} and \Cref{tab2}, using raw ground truth and improved ground truth, respectively. It is obvious that our method with D-HRNet can achieve state-of-the-art performance. Besides, our Mono-ViFI brings notable improvements to the three depth architectures without increasing model complexity (see each subsection separated by dotted lines). Despite that the difference between Mono-ViFI (D-HRNet) and RA-Depth~\cite{radepth} under single-frame mode is slight in \Cref{tab1}, it is more significant in \Cref{tab2} when using improved ground truth. For multi-frame inference, Mono-ViFI with ResNet18 has already shown advantages over ManyDepth~\cite{manydepth} and DynamicDepth~\cite{dynamicdepth} on some metrics including Sq Rel, RMSE, RMSE log, $\delta_2$ and $\delta_3$. When changing to D-HRNet backbone, proposed Mono-ViFI can totally outperform them. Although our models have more multiply-add computations (MACs) when predicting for a single frame, the average MACs per frame are comparable when processing a long video sequence. \cref{fig:vis} depicts two visual comparison results on KITTI, where we choose D-HRNet as the baseline depth model. It can be seen that depth maps look qualitatively similar among all four methods. However, the error maps reveal mistakes of D-HRNet baseline and RA-Depth in some ambiguous regions like bushes (see the red boxes). In contrast, our Mono-ViFI have dramaticly lower error in these regions, especially under multi-frame reasoning. For the quantitative outcomes on KITTI under high resolution of $1024 \times 320$ and more visualization results, please refer to our supplementary.

\begin{figure*}[t]
	\vspace{-8mm}
	\centering
	\begin{minipage}{0.49\linewidth}
        \begin{table}[H]
    \begin{center}
    \caption{Performance comparison on KITTI~\cite{kitti} using improved ground truth from~\cite{im-gt}, with a resolution of $640 \times 192$.}
    \vspace{-6mm}
    \label{tab2}
    \resizebox{1.0\textwidth}{!}{
    \begin{tabular}{|c||c|c|c|c|c|c|c|}
    \hline
    Method & \cellcolor{col1}Abs Rel & \cellcolor{col1}Sq Rel & \cellcolor{col1}RMSE  & \cellcolor{col1}RMSE log & \cellcolor{col2}$\delta_1$ & \cellcolor{col2}$\delta_2$ & \cellcolor{col2}$\delta_3$\\
    \hline
    \rowcolor{gray!20} \multicolumn{8}{|c|}{Single-frame Inference} \\
    \hline
    PackNet-SfM~\cite{packnet} & 0.078 & 0.420 & 3.485 & 0.121 & 0.931 & 0.986 &  0.996 \\
    R-MSFM6~\cite{r-msfm} & 0.088 & 0.492 & 3.837 & 0.135 & 0.915 & 0.983 & 0.995 \\
    DIFFNet~\cite{diffnet} & 0.076 & 0.414 & 3.495 & 0.119 & 0.936 & 0.988 & 0.996 \\
    MonoViT~\cite{monovit} & \underline{0.074} & 0.388 & 3.414 & 0.115 & 0.938 & 0.989 & \underline{0.997} \\
    \hdashline 
    Monodepth2 (ResNet18)~\cite{mono2} & 0.090 & 0.545 & 3.942 & 0.137 & 0.914 & 0.983 & 0.995\\
    \textbf{Mono-ViFI (ResNet18)} &0.084 & 0.431 & 3.618 & 0.126 & 0.924 & 0.986 & \underline{0.997} \\   
    \hdashline 
    Lite-Mono~\cite{lite-mono} & 0.082 & 0.455 & 3.685 & 0.127 & 0.923 & 0.985 & 0.996 \\
    \textbf{Mono-ViFI (Lite-Mono)} & 0.080 & 0.400 & 3.497 & 0.121 & 0.930 & 0.987 & \underline{0.997} \\    
    \hdashline 
    D-HRNet~\cite{radepth} & 0.077  & 0.423 & 3.496 & 0.119 & 0.935  & 0.987 & 0.996\\
    RA-Depth (D-HRNet)~\cite{radepth} & \underline{0.074} & \underline{0.363} & \underline{3.349} & \underline{0.114} & \underline{0.940} & \underline{0.990} & \underline{0.997} \\
    \textbf{Mono-ViFI (D-HRNet)} & \textbf{0.072}  & \textbf{0.342} & \textbf{3.254} & \textbf{0.111} & \textbf{0.941}  & \textbf{0.991} & \textbf{0.998}\\
    \hline 
    \rowcolor{gray!20} \multicolumn{8}{|c|}{Multi-frame Inference} \\
    \hline
    ManyDepth (ResNet18)~\cite{manydepth} & 0.070 & 0.399 & 3.455 & 0.113 & 0.941 & 0.989 & \underline{0.997}\\
    MOVEDepth~\cite{movedepth} & \textbf{0.065} & 0.377 & 3.449 & \underline{0.112} & \underline{0.942} & 0.988 & 0.996 \\       
    \textbf{Mono-ViFI (ResNet18)} & 0.074 & 0.374 & 3.381 & 0.113 & 0.941 & \underline{0.990} & \underline{0.997}\\
    \textbf{Mono-ViFI (Lite-Mono)} & 0.074 & \underline{0.354} & \underline{3.372} & 0.114 & 0.939 & 0.989& \underline{0.997}\\
    \textbf{Mono-ViFI (D-HRNet)} & \underline{0.066} & \textbf{0.294} & \textbf{3.099} & \textbf{0.102} & \textbf{0.952} & \textbf{0.992} & \textbf{0.998}\\
    \hline 
    \end{tabular}}
    % \vspace{-5mm}
    \end{center}
    \end{table}

	\end{minipage}
	\hspace{0.00\linewidth}
	\begin{minipage}{0.49\linewidth}
        \begin{table}[H]
    \begin{center}
    \caption{Performance comparison on Cityscapes~\cite{city} with a resolution of $416 \times 128 $ unless stated.}
    \vspace{-3mm}
    \label{tab3}
    \resizebox{0.99\textwidth}{!}{
    \begin{tabular}{|c||c|c|c|c|c|c|c|}
    \hline
    Method & \cellcolor{col1}Abs Rel & \cellcolor{col1}Sq Rel & \cellcolor{col1}RMSE  & \cellcolor{col1}RMSE log & \cellcolor{col2}$\delta_1$ & \cellcolor{col2}$\delta_2$ & \cellcolor{col2}$\delta_3$\\
    \hline
    \rowcolor{gray!20} \multicolumn{8}{|c|}{Single-frame Inference} \\
    \hline
    Videos in the wild~\cite{wild} & 0.127 & 1.330 & 6.960 & 0.195 & 0.830 & 0.947 & 0.981 \\
    Li \etal~\cite{li} & 0.119 & 1.290 & 6.980 & 0.190 & 0.846 & 0.952 & 0.982 \\
    InstaDM~\cite{iapc} ($832 \times 256$) & 0.111 & 1.158 & 6.437 & 0.182 & 0.868 & 0.961 & 0.983 \\
    SD-SSMDE~\cite{sd-ssmde} & 0.110 & 0.988 & 5.953 & 0.165 & - & - & - \\
    \hdashline 
    Monodepth2 (ResNet18)~\cite{mono2}& 0.129 & 1.569 & 6.876 & 0.187 & 0.849 & 0.957 & 0.983\\
    \textbf{Mono-ViFI (ResNet18)}& 0.112 & 1.003 & 5.899 & 0.165 & 0.872 & 0.966 & 0.990 \\   
    \hdashline 
    Lite-Mono~\cite{lite-mono} & 0.114 & 1.194 & 6.328 & 0.172 & 0.871 & 0.967 & 0.989 \\
    \textbf{Mono-ViFI (Lite-Mono)} & \underline{0.107} & \underline{0.940} & \underline{5.794} & \underline{0.159} & 0.876 & \underline{0.968} & \underline{0.991} \\    
    \hdashline 
    D-HRNet~\cite{radepth} & 0.111  & 1.142 & 6.203 & 0.169 & \underline{0.879} & \underline{0.968} & 0.989\\
    \textbf{Mono-ViFI (D-HRNet)} & \textbf{0.105}  & \textbf{0.880} & \textbf{5.550} & \textbf{0.156} & \textbf{0.881} & \textbf{0.969} & \textbf{0.992}\\
    \hline 
    \rowcolor{gray!20} \multicolumn{8}{|c|}{Multi-frame Inference} \\
    \hline
    ManyDepth (ResNet18)~\cite{manydepth} & 0.114 & 1.193 & 6.223 & 0.170 & 0.875 & 0.967 & 0.989\\
    DynamicDepth~\cite{dynamicdepth} & \underline{0.104} & 1.011 & 5.987 & 0.159 & \textbf{0.889} & \textbf{0.972} & \underline{0.991} \\
    TC-Depth~\cite{tcdepth} & 0.110 & 0.958 & 5.820 & - & 0.867 & - & \underline{0.991} \\
    \textbf{Mono-ViFI (ResNet18)} & 0.107 & 0.934 & 5.688 & 0.159 & 0.880 & 0.969 & 0.990\\
    \textbf{Mono-ViFI (Lite-Mono)} & 0.105 & \underline{0.904} & \underline{5.646} & \underline{0.157} & 0.880 & 0.970 & \underline{0.991}\\
    \textbf{Mono-ViFI (D-HRNet)} & \textbf{0.102} & \textbf{0.836} & \textbf{5.395} & \textbf{0.153} & \underline{0.885} & \underline{0.971} & \textbf{0.992}\\
    \hline 
    \end{tabular}}
    % \vspace{-5mm}
    \end{center}
    \end{table}

	\end{minipage}
	\vspace{-5mm}
\end{figure*}

\begin{table}[t]
    \begin{center}
    \caption{Evaluation results on dynamic objects and static regions of Cityscapes~\cite{city} dataset under $416 \times 128 $ resolution. Note that DynamicDepth relies on depth prior and segmentation masks of dynamic objects when reasoning.}
    \vspace{-3mm}
    \label{tab4}
    \resizebox{0.85\textwidth}{!}{
    \begin{tabular}{|c|c||c|c|c|c|c|c|c|}
    \hline
    Method & Test Area & \cellcolor{col1}Abs Rel & \cellcolor{col1}Sq Rel & \cellcolor{col1}RMSE  & \cellcolor{col1}RMSE log & \cellcolor{col2}$\delta_1$ & \cellcolor{col2}$\delta_2$ & \cellcolor{col2}$\delta_3$\\
    \hline
    \rowcolor{gray!20} \multicolumn{9}{|c|}{Multi-frame Inference} \\
    \hline
    ManyDepth (ResNet18)~\cite{manydepth} & \multirow{3}{*}{Dynamic Objects} & 0.172 & 2.206 & 5.924 & 0.216 & 0.792 & 0.917 & 0.967 \\
    DynamicDepth~\cite{dynamicdepth} & & \textbf{0.144} & 1.554 & 5.238 & \textbf{0.185} & \textbf{0.831} & \textbf{0.950} & \textbf{0.982} \\
    \textbf{Mono-ViFI (ResNet18)} & & 0.162 & \textbf{1.453} & \textbf{5.114} & 0.199 & 0.794 & 0.924 & 0.977 \\
    \hline
    ManyDepth (ResNet18)~\cite{manydepth} & \multirow{3}{*}{Static Regions} & 0.107 & 1.090 & 6.314 & 0.160 & 0.882 & 0.972 & 0.991 \\
    DynamicDepth~\cite{dynamicdepth} & & 0.100 & 0.977 & 6.155 & 0.155 & \textbf{0.892} & 0.974 & \textbf{0.992} \\
    \textbf{Mono-ViFI (ResNet18)} & & \textbf{0.099} & \textbf{0.907} & \textbf{5.835} & \textbf{0.149} & 0.890 & \textbf{0.975} & \textbf{0.992} \\
    \hline 
    \end{tabular}}
    \vspace{-9mm}
    \end{center}
    \end{table}

\noindent \textbf{Cityscapes Results.} Results of Cityscapes in \Cref{tab3} demonstrate the same trend as KITTI. Moreover, proposed multi-frame model with ResNet18 can beat ManyDepth on all metrics, while our Mono-ViFI is better than DynamicDepth only on partial indexes. It is because Cityscapes with more dynamic scenes is more challenging than KITTI, and DynamicDepth leverages additional information to cope with dynamic objects. For fine-grained analysis, we separately evaluate on the dynamic objects and static regions, using the segmentation masks of dynamic objects provided in \cite{dynamicdepth}, as listed in \Cref{tab4}. On dynamic objects, Mono-ViFI with ResNet18 surpasses ManyDepth by a large margin, demonstrating that our multi-frame inference paradigm can sidestep the issue of dynamic objects to a greater extent compared to ManyDepth. DynamicDepth utilizes depth prior and segmentation masks of dynamic objects to tackle this problem for multi-frame inference. Even though, our method still excels it on Sq Rel and RMSE. When evaluating on static regions, proposed Mono-ViFI shows absolute superiority over ManyDepth and DynamicDepth.

\noindent \textbf{Generalization Results.} We evaluate the KITTI models on Cityscapes and Make3D without fine-tuning, as depicted in \Cref{tab5} and \Cref{tab6}. For qualitative comparison, we give a visualization example in \cref{fig4}, where KITTI models are used to predict the multi-frame depth of a frame in a YouTube video, downloaded from the Wind Walk Travel Videos channel. These results indicate that our Mono-ViFI has better generalization ability than existing methods.

%  Comparison results of Cityscapes listed in Table~\ref{tab2} also show the same trend. To validate the generalization ability across datasets, we use the models trained on KITTI to evaluate on Cityscapes without fine-tuning. The outcomes are depicted in Table~\ref{tab3}, which demonstrate that our Mono-ViFI can also improve the generalization capability of existing models. For fine-grained analysis, we separately evaluate on the dynamic objects and static regions, using the segmentation masks of dynamic objects provided in \cite{dynamicdepth}.

% Note that we carry out KITTI comparison with raw ground truth under two different resolutions and put the generalization to Make3D in our supplementary.

\begin{figure*}[t]
	\vspace{-7mm}
	\centering
	\begin{minipage}{0.49\linewidth}
        \begin{table}[H]
    \begin{center}
        %\vspace{-5mm}
    \caption{Generalization comparison results from KITTI~\cite{kitti} to Cityscapes~\cite{city}.}
    \label{tab5}
    \vspace{-3mm}
    \renewcommand{\arraystretch}{1.1}{
    \resizebox{0.9\textwidth}{!}{\setlength{\tabcolsep}{0.8mm}{
    \begin{tabular}{|c||c|c|c|c|}
    \hline
    Method  & \cellcolor{col1}Abs Rel & \cellcolor{col1}Sq Rel & \cellcolor{col1}RMSE  & \cellcolor{col1}RMSE log \\
    \hline
    \rowcolor{gray!20} \multicolumn{5}{|c|}{Single-frame Inference} \\
    \hline
    Monodepth2 (ResNet18)~\cite{mono2} & 0.164&1.890&8.985& 0.242\\ 
    \textbf{Mono-ViFI (ResNet18)} & 0.156 & 1.603 & \underline{8.114} & 0.225\\
    \hdashline
    Lite-Mono~\cite{lite-mono} & 0.153 & 1.642 & 8.409 & 0.226 \\
    \textbf{Mono-ViFI (Lite-Mono)} & \underline{0.149} & \underline{1.562} & 8.163 & \underline{0.221} \\
    \hdashline
    D-HRNet~\cite{radepth} & 0.150 & 1.663 & 8.458 & 0.224 \\
    \textbf{Mono-ViFI (D-HRNet)} & \textbf{0.134}  & \textbf{1.360} & \textbf{7.581} & \textbf{0.201} \\
    \hline
    \rowcolor{gray!20} \multicolumn{5}{|c|}{Multi-frame Inference} \\
    \hline
    ManyDepth (ResNet18)~\cite{manydepth} & 0.170 & 1.789 & 8.357 & 0.236 \\
    MOVEDepth~\cite{movedepth} & 0.164 & 1.780 & 8.678 & 0.238 \\
    \textbf{Mono-ViFI (ResNet18)} & 0.150 & 1.502 & \underline{7.895} & 0.217 \\
    \textbf{Mono-ViFI (Lite-Mono)} & \underline{0.145} & \underline{1.455} & 7.986 & \underline{0.215} \\
    \textbf{Mono-ViFI (D-HRNet)} & \textbf{0.132} & \textbf{1.283} & \textbf{7.384} & \textbf{0.197} \\
    \hline
    \end{tabular}}}}
    % \vspace{-3mm}
    \end{center}
    \end{table}

	\end{minipage}
	\hspace{0.00\linewidth}
	\begin{minipage}{0.49\linewidth}
        \begin{table}[H]
    \begin{center}
    \caption{Generalization comparison results from KITTI~\cite{kitti} to Make3D~\cite{make3d}.}
    \label{tab6}
    \vspace{-3mm}
    \renewcommand{\arraystretch}{1.1}{
    \resizebox{0.9\textwidth}{!}{\setlength{\tabcolsep}{0.8mm}{
    \begin{tabular}{|c|c|c|c|c|}
    \hline
    Method & \cellcolor{col1}Abs Rel & \cellcolor{col1}Sq Rel & \cellcolor{col1}RMSE  & \cellcolor{col1}RMSE $\rm log_{10}$ \\
    \hline
    \rowcolor{gray!20} \multicolumn{5}{|c|}{Single-frame Inference} \\
    \hline
    HR-Depth~\cite{hrdepth} & 0.315 & 3.208 & 7.024 & 0.159\\
    R-MSFM6~\cite{r-msfm} & 0.334 & 3.285 & 7.212 & 0.169\\
    DIFFNet~\cite{diffnet} & 0.309 & 3.313 & 7.008 & 0.155\\
    Lite-Mono-8M~\cite{lite-mono} & 0.305 & 3.060 & 6.981 & 0.158 \\
    BRNet~\cite{brnet} & 0.302 & 3.133 & 7.068 & 0.156 \\
    DaCCN~\cite{daccn} & \underline{0.290} & \underline{2.873} & \underline{6.656} & \underline{0.149} \\
    \hdashline
    Monodepth2 (ResNet18)~\cite{mono2}  & 0.322 & 3.589 & 7.417 & 0.163 \\
    \textbf{Mono-ViFI (ResNet18)}  & 0.294 & 3.081 & 6.864 & 0.151\\
    \hdashline
    Lite-Mono~\cite{lite-mono}  & 0.318 & 3.412 & 7.255 & 0.161 \\
    \textbf{Mono-ViFI (Lite-Mono)} & 0.302 & 3.286 & 7.154 & 0.154 \\
    \hdashline
    D-HRNet~\cite{radepth}  & 0.308 & 3.488 & 7.246 & 0.156 \\
    \textbf{Mono-ViFI (D-HRNet)} & \textbf{0.280}  & \textbf{2.727} & \textbf{6.596} & \textbf{0.146} \\
    \hline
    \end{tabular}}}}
    \end{center}
    \end{table}
	\end{minipage}
	\vspace{-6mm}
\end{figure*}

\begin{figure*}[t]
    \centering
    %   \fbox{\rule{0pt}{2in} \rule{0.9\linewidth}{0pt}}
    \includegraphics[width=0.88\linewidth]{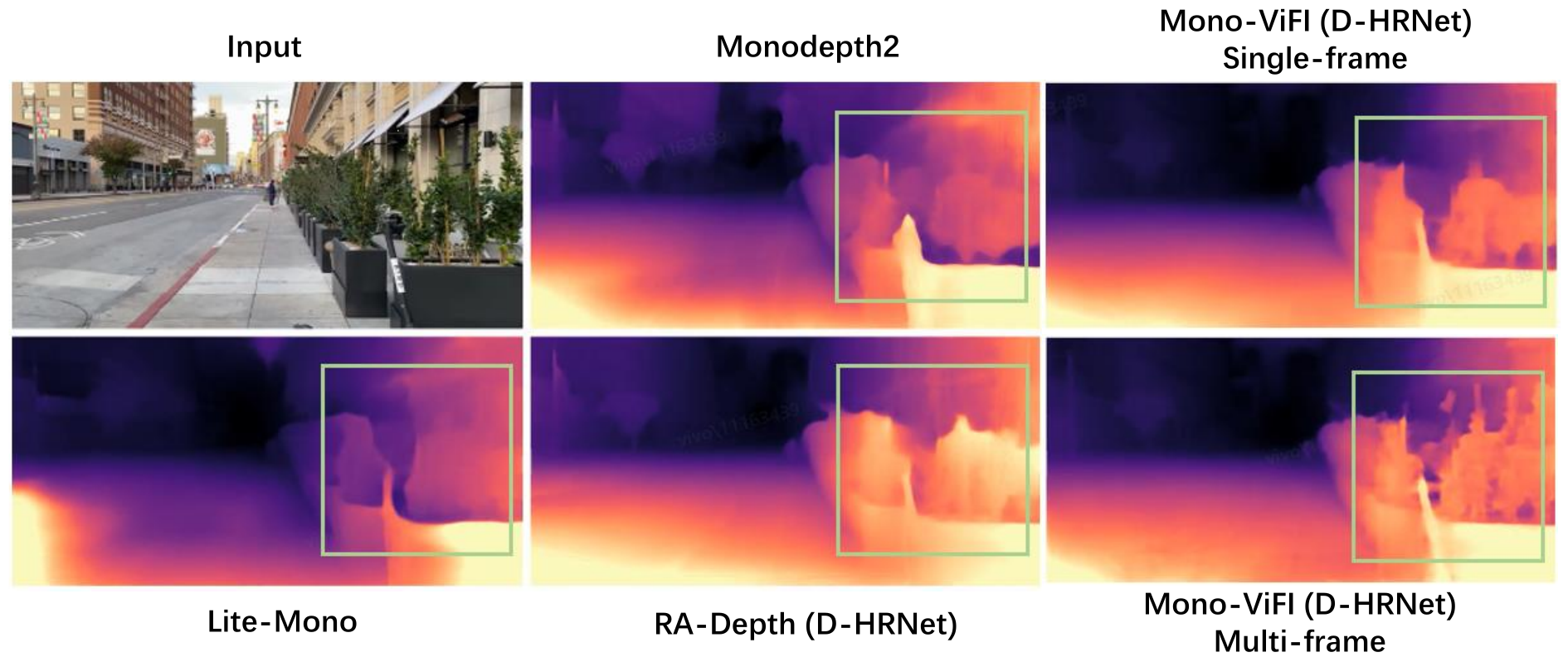}
    \vspace{-3mm}
    \caption{Qualitative generalization results of a YouTube video example. All depth maps are predicted under $640\times360$ resolution using KITTI models trained under $640\times192$.}
    \label{fig4}
    \vspace{-6mm}
    \end{figure*}

\begin{table}[t]
    \begin{center}
    \caption{Ablation results of each proposed component on Cityscapes~\cite{city}. SA is spatial augmentation. TA is temporal augmentation. SVDC denotes standard view depth consistency. SADC denotes scale-aware depth consistency. FA is feature alignment. MAFA is motion-aware feature alignment. OAFF means occlusion-alleviated feature fusion.}
    \label{tab7}
    \vspace{-3mm}
    \renewcommand{\arraystretch}{1.1}{
    \resizebox{0.85\textwidth}{!}{
    \begin{tabular}{|c||c|c|c|c|c|c|c||c|c|c|c|}
    \hline
    Method & SA & TA & SVDC & SADC & FA & MAFA & OAFF & \cellcolor{col1}Abs Rel & \cellcolor{col1}Sq Rel & \cellcolor{col1}RMSE & \cellcolor{col2}$\delta_1$ \\
    \hline
    \rowcolor{gray!20} \multicolumn{12}{|c|}{Single-frame Inference} \\
    \hline
    Baseline (Monodepth2~\cite{mono2}) & & & & & & & & 0.129 & 1.569 & 6.876 & 0.849 \\
    Baseline + TA & & \checkmark & & & & & & 0.119 & 1.254 & 6.523 & 0.857 \\
    Baseline + SVDC & &  &\checkmark & & & & & 0.117 & 1.186 & 6.492 & 0.862 \\
    Baseline + SA & \checkmark & & & & & & & 0.122 & 1.386 & 6.557 & 0.856 \\
    Baseline + SA + SADC & \checkmark & & & \checkmark & & & & 0.116 & 1.079 & 6.189 & 0.868 \\
    \textbf{Mono-ViFI (Single-frame)} & \checkmark& \checkmark &\checkmark & \checkmark& & & & \textbf{0.112} & \textbf{1.003} & \textbf{5.899} & \textbf{0.872} \\
    \hline
    \rowcolor{gray!20} \multicolumn{12}{|c|}{Multi-frame Inference} \\
    \hline
    Baseline (direct feature fusion)& & & & & & & & 0.120 & 1.360 & 6.491 & 0.869 \\
    FA (not motion-aware) & & & & & \checkmark & & & 0.116 & 1.217 & 6.387 & 0.871 \\
    MAFA & & & & & &\checkmark & & 0.114 & 1.150 & 6.115 & 0.874 \\
    MAFA + OAFF & & & & & &\checkmark & \checkmark& 0.111 & 1.107 & 6.012 & 0.876 \\    
    MAFA + OAFF + TA & &\checkmark & & & &\checkmark & \checkmark& 0.109 & 1.023 & 5.892 & 0.877 \\    
    \textbf{Mono-ViFI (Multi-frame)} & \checkmark&\checkmark &\checkmark & \checkmark& &\checkmark & \checkmark& \textbf{0.107} & \textbf{0.934} & \textbf{5.688} & \textbf{0.880} \\ 
    \hline
    \end{tabular}}}
    \vspace{-4mm}
    \end{center}
    \end{table}

\subsection{Ablation Study}
In this part, we explore the influence of each component in Mono-ViFI. Our ablations are conducted on Cityscapes or KITTI with ResNet18 backbone.  

\noindent \textbf{Single-frame Ablation.} We first ablate on the single-frame model. As listed in the top half part of \Cref{tab7}, the standard view depth consistency and spatial augmentation can bring gains individually. Note that scale-aware depth consistency $L_{sa}$ should be based on spatial augmentation. When adding $L_{sa}$ to spatial augmentation, the accuracy can be further improved. And this leads to the most significant enhancement among all components, since $L_{sa}$ can enforce the model to learn scale relationship from scene variations. In orthogonal dimension, our proposed temporal augmentation can also reduce the prediction error by providing more guidance from newly synthesized camera views. When integrating all above elements, our single-frame model can achieve the best performance.

\noindent \textbf{Multi-frame Ablation.} To probe the effect of each component in our VFI-assisted multi-frame fusion module, we conduct ablation experiments under multi-frame inference, whose results are summarized in the bottom half part of \Cref{tab7}. The baseline is stacking the raw feature of three frames directly, \ie, direct feature fusion. Then, we gradually apply components of the proposed fusion module, including motion-aware feature alignment and occlusion-alleviated feature fusion. The results demonstrate that these strategies can bring cumulative improvements. Furthermore, when incorporating spatial augmentation, temporal augmentation and relevant depth consistency losses, our multi-frame model can be further promoted like the single-frame setting.

\begin{table*}[t]
    \begin{center}
        %\vspace{-5mm}
    \caption{Ablation of different flow-based VFI models on KITTI~\cite{kitti}. }
    \label{tab8}
    \vspace{-3mm}
    \renewcommand{\arraystretch}{1.1}{
    \resizebox{0.6\textwidth}{!}{
    \begin{tabular}{|c||c|c|c|c|c|c|c|}
    \hline
    VFI Model & \cellcolor{col1}Abs Rel & \cellcolor{col1}Sq Rel & \cellcolor{col1}RMSE & \cellcolor{col1}RMSE log & \cellcolor{col2}$\delta_1$ & \cellcolor{col2}$\delta_2$ & \cellcolor{col2}$\delta_3$\\
    \hline
    \rowcolor{gray!20} \multicolumn{8}{|c|}{Multi-frame Inference} \\
    \hline
    IFRNet~\cite{ifrnet} (Ours) & 0.099 & 0.661 & 4.321 & 0.174 & 0.898 & 0.966 & 0.985 \\
    RIFE~\cite{rife} & 0.100 & 0.672 & 4.352 & 0.175 & 0.896 & 0.965 & 0.985 \\
    AMT~\cite{amt} & 0.099 & 0.653 & 4.335 & 0.174 & 0.897 & 0.966 & 0.985 \\
    \hline
    \end{tabular}}}
    \vspace{-9mm}
    \end{center}
    \end{table*}

\noindent \textbf{Ablation of different VFI models.} Apart from IFRNet~\cite{ifrnet}, we additionally experiment with another two flow-based VFI models, RIFE~\cite{rife} and AMT~\cite{amt}. The results are  depicted in \Cref{tab8}. It can be observed that there is only a negligible performance difference among the three VFI networks, which reveals that our method is robust to different VFI models.

\section{Conclusion}
In conclusion, this paper presents a novel self-supervised learning framework, named Mono-ViFI, to bilaterally connect single- and multi-frame depth estimation in a unified manner. First, a flow-based VFI model is employed for temporal augmentation. Then, we propose a VFI-assisted multi-frame fusion module for more accurate multi-frame depth. Besides, resizing-cropping is extended to affine transformation as spatial augmentation for enriching data diversity. Finally, we develop a new triplet depth consistency loss for better regularization. Experiments show that our approach can unleash the potential of existing depth models to a greater extent, but without increasing inference complexity.

\appendix

\title{Supplementary Material for Mono-ViFI} 
\titlerunning{Mono-ViFI: A Unified Framework for Self-supervised Monocular Depth}
\author{Jinfeng Liu \and
Lingtong Kong\textsuperscript{\Letter} \and
Bo Li \and Zerong Wang \and Hong Gu \and Jinwei Chen}
\authorrunning{J.~Liu et al.}
\institute{vivo Mobile Communication Co., Ltd, China\\
\email{\{liujinfeng,ltkong,libra,wangzerong,guhong,jinwei.chen\}@vivo.com}}
\maketitle

\section{Additional Method Details}

\subsection{Affine Transformation}
For an image $I$, suppose it is rotated by an angle of $\theta$, where counterclockwise is the positive direction. Then, resizing-cropping is applied to generate the final augmented image $\widetilde{I}$. Let $(c_x, c_y)$ and $(p_x, p_y)$ denote the image center and the cropping center respectively, $f_s \ge 1$ denote the resizing factor, as shown in \cref{fig:affine}. Hence, for a pixel $(x,y)$ and its depth $D$ in the original image $I$, the corresponding pixel $(\widetilde{x},\widetilde{y})$ and depth $\widetilde{D}$ in $\widetilde{I}$ can be obtained by:
\begin{equation}
    \label{sup:eq1}
    \begin{bmatrix} \widetilde{x} \\ \widetilde{y} \\ 1 
    \end{bmatrix}=\boldsymbol{R}\begin{bmatrix} f_s(x-c_x) \\ f_s(y-c_y) \\ 1 
    \end{bmatrix} + 
    \begin{bmatrix} f_s(c_x-p_x)+c_x \\ f_s(c_y-p_y)+c_y \\ 0
    \end{bmatrix},
\end{equation}

\begin{equation}
    \label{sup:eq2}
    \widetilde{D} = D/f_s,
\end{equation}
where $\boldsymbol{R}$ is the rotation transformation matrix:
\begin{equation}
    \boldsymbol{R}=\begin{bmatrix} \cos\theta & \sin\theta & 0 \\
        -\sin\theta & \cos\theta & 0 \\
        0 & 0 & 1
    \end{bmatrix}. 
\end{equation}
\cref{sup:eq2} is due to the assumption~\cite{planedepth,bdedepth} mentioned in main paper that the depth of an object in $I_t$ is $f_s$ times of that in $\widetilde{I}_t$. And \cref{sup:eq1} can be rewritten as:
\begin{small} 
\begin{equation}
    \label{sup:eq4}
\begin{aligned}
    \begin{bmatrix} \widetilde{x} \\ \widetilde{y} \\ 1 
    \end{bmatrix} 
    &= \boldsymbol{R}\begin{bmatrix} f_s x \\ f_s y \\ f_s
    \end{bmatrix} + 
    \boldsymbol{R}\begin{bmatrix} -f_s c_x \\ -f_s c_y \\ 1-f_s
    \end{bmatrix} + 
    \begin{bmatrix} f_s(c_x-p_x)+c_x \\ f_s(c_y-p_y)+c_y \\ 0    
    \end{bmatrix} \\
    &= f_s \boldsymbol{R}\begin{bmatrix} x \\ y \\ 1
    \end{bmatrix} + f_s \boldsymbol{q}, 
\end{aligned}
\end{equation}
\end{small}
where $\boldsymbol{q}$ can be denoted as:
\begin{figure}[t]
    \centering
    %   \fbox{\rule{0pt}{2in} \rule{0.9\linewidth}{0pt}}
    \includegraphics[width=0.7\linewidth]{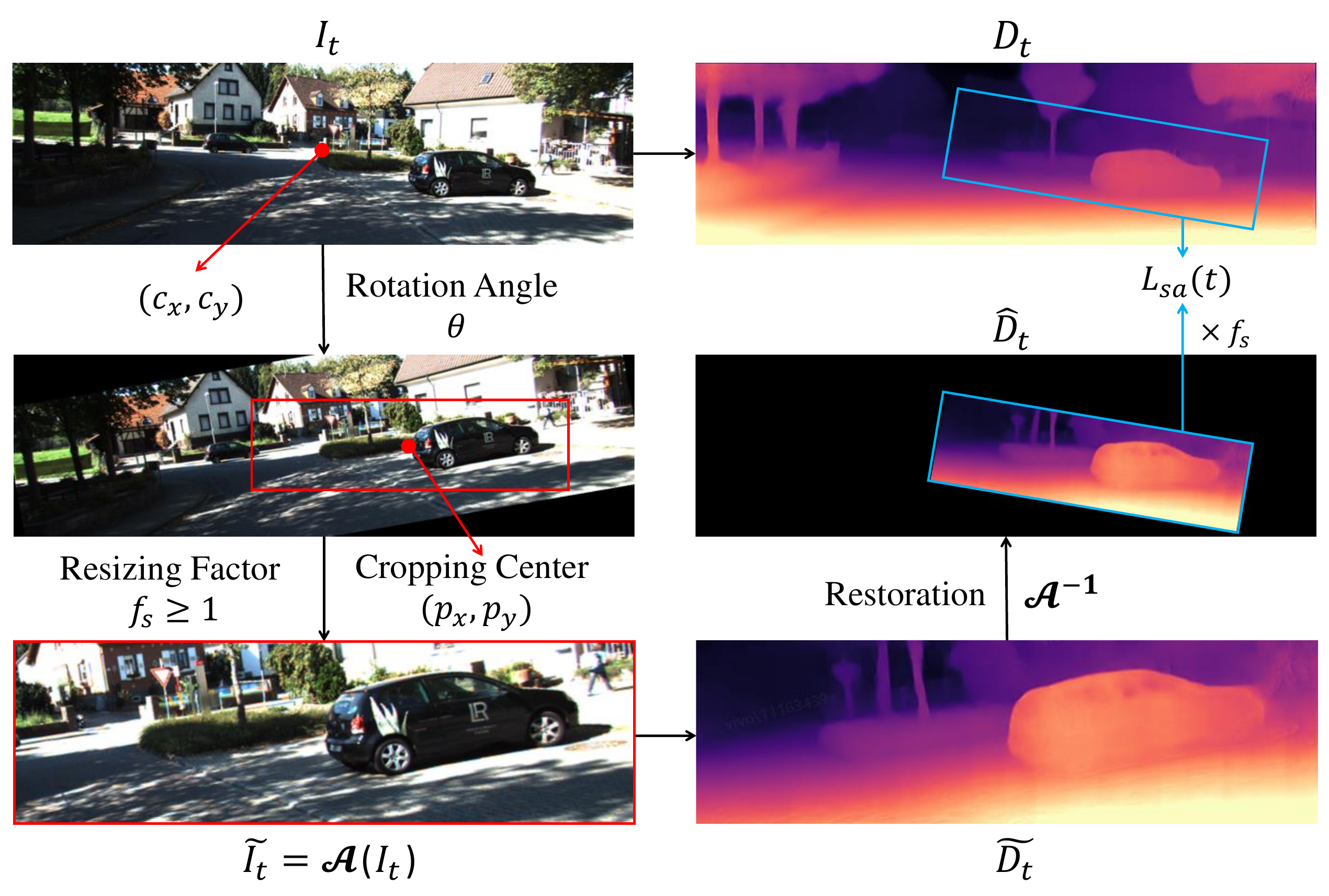}
    \caption{Affine Transformation.}
    \label{fig:affine}
    \vspace{-5mm}
\end{figure}
\begin{equation}
    \boldsymbol{q} = \boldsymbol{R}\begin{bmatrix} -c_x \\ -c_y \\ 1/f_s-1
    \end{bmatrix} + 
    \begin{bmatrix} (c_x-p_x)+c_x/f_s \\ (c_y-p_y)+c_y/f_s \\ 0    
    \end{bmatrix}.
\end{equation}
The 3D world coordinates of $(x,y)$ in the original image $I$ and $(\widetilde{x},\widetilde{y})$ in the augmented image $\widetilde{I}$, denoted as $\boldsymbol{P}$ and $\boldsymbol{\widetilde{P}}$, respectively, can be expressed as:
\begin{equation}
    \label{sup:eq6}
    \boldsymbol{P}=\begin{bmatrix} X \\ Y \\ D 
    \end{bmatrix}=D\boldsymbol{K}^{-1}\begin{bmatrix} x \\ y \\ 1 
    \end{bmatrix},   ~~~~~~~   
    \boldsymbol{\widetilde{P}}=\widetilde{D}\boldsymbol{K}^{-1}\begin{bmatrix} \widetilde{x} \\ \widetilde{y} \\ 1 
    \end{bmatrix}. 
\end{equation}
Combining \cref{sup:eq2,sup:eq4,sup:eq6}, we derive the relationship between $\boldsymbol{P}$ and $\boldsymbol{\widetilde{P}}$:
\begin{equation}
\begin{aligned}
    \boldsymbol{\widetilde{P}} &=\widetilde{D}\boldsymbol{K}^{-1}\begin{bmatrix} \widetilde{x} \\ \widetilde{y} \\ 1 
    \end{bmatrix} \\
    &=\frac{1}{f_s} D\boldsymbol{K}^{-1}(f_s \boldsymbol{R}\begin{bmatrix} x \\ y \\ 1
    \end{bmatrix} + f_s \boldsymbol{q}) \\
    &=D\boldsymbol{K}^{-1}(\frac{1}{D}\boldsymbol{RKP}+\boldsymbol{q}) \\
    &=\boldsymbol{K^{-1}RKP} + D\boldsymbol{K}^{-1}\boldsymbol{q} \\
    &=\boldsymbol{K^{-1}RK}\begin{bmatrix} X \\ Y \\ D 
    \end{bmatrix} + \boldsymbol{K}^{-1}
    \begin{bmatrix} \boldsymbol{0} & \boldsymbol{0} & \boldsymbol{q} \\
    \end{bmatrix} \begin{bmatrix} X \\ Y \\ D 
    \end{bmatrix} \\
    &=(\boldsymbol{K^{-1}RK}+\boldsymbol{K}^{-1}
    \begin{bmatrix} \boldsymbol{0} & \boldsymbol{0} & \boldsymbol{q}  \\
    \end{bmatrix})\boldsymbol{P},
\end{aligned}
\end{equation}
where $\boldsymbol{0}=[0~~0~~0]^T$ is the zero column vector. Therefore, the transformation $\boldsymbol{R}_c$ from $\boldsymbol{P}$ to $\boldsymbol{\widetilde{P}}$ can be formulated as:
\begin{equation}
    \boldsymbol{R}_c = \boldsymbol{K}^{-1}\boldsymbol{R}\boldsymbol{K}+\boldsymbol{K}^{-1}\begin{bmatrix} \boldsymbol{0} & \boldsymbol{0} & \boldsymbol{q}  \\
    \end{bmatrix},
\end{equation}
where $\boldsymbol{R}_c$ is determined only by the camera intrinsics and the augmentation parameters. The affine transformation $\mathcal{A}$ is applied to both the target image $I_t$ and the source image $I_s$. Consequently, we can add subscripts to $\{\boldsymbol{P},\boldsymbol{\widetilde{P}}\}$, generating $\{\boldsymbol{P}_t,\boldsymbol{\widetilde{P}}_t\}$ for $I_t$ and $\{\boldsymbol{P}_s,\boldsymbol{\widetilde{P}}_s\}$ for $I_s$. Their relationships can be written as:
\begin{equation} \label{sup:eq9}
\begin{aligned}
    \boldsymbol{\widetilde{P}}_t&=\boldsymbol{R}_c\boldsymbol{P}_t, \\
    \boldsymbol{\widetilde{P}}_s&=\boldsymbol{R}_c\boldsymbol{P}_s, \\
    \boldsymbol{P}_s&=\boldsymbol{R}_{t \rightarrow s}\boldsymbol{P}_t + \boldsymbol{t}_{t \rightarrow s},
\end{aligned}
\end{equation}
where $\boldsymbol{R}_{t \rightarrow s}$ and $\boldsymbol{t}_{t \rightarrow s}$ represent the camera rotation and translation from $I_t$ to $I_s$. From \cref{sup:eq9}, we can further derive the relationship between $\boldsymbol{\widetilde{P}}_t$ and $\boldsymbol{\widetilde{P}}_s$ as:
\begin{equation}
    \boldsymbol{\widetilde{P}}_s=\boldsymbol{R}_c\boldsymbol{R}_{t \rightarrow s}\boldsymbol{R}_c^{-1}\boldsymbol{\widetilde{P}}_t+\boldsymbol{R}_c\boldsymbol{t}_{t \rightarrow s}.
\end{equation}
Thus, the rectified camera pose $\boldsymbol{\widetilde{T}}_{t \rightarrow s}=[\boldsymbol{\widetilde{R}}_{t \rightarrow s}|\boldsymbol{\widetilde{t}}_{t \rightarrow s}]$ from the augmented target image to the augmented source image can be denoted as:
\begin{equation}
    \boldsymbol{\widetilde{R}}_{t \rightarrow s}=\boldsymbol{R}_c\boldsymbol{R}_{t \rightarrow s}\boldsymbol{R}_c^{-1},~~~~\boldsymbol{\widetilde{t}}_{t \rightarrow s}=\boldsymbol{R}_c\boldsymbol{t}_{t \rightarrow s}.
\end{equation}
For the parameters of affine transformation, we randomly sample the resizing factor $f_s$ from $[1.2, 2.0]$ and the rotation angle $\theta$ from $[-5^{\circ},5^{\circ}]$.

\begin{table*}[t]
    \begin{center}
    \caption{Quantitative performance comparison results on KITTI~\cite{kitti} benchmark using the raw ground truth, with the high resolution of $1024 \times 320$.}
    \vspace{-1mm}
    \label{sup:tab1}
    \resizebox{0.95\textwidth}{!}{
    \begin{tabular}{|c|c||c|c|c|c|c|c|c|}
    \hline
    Method  & Test frames & \cellcolor{col1}Abs Rel & \cellcolor{col1}Sq Rel & \cellcolor{col1}RMSE  & \cellcolor{col1}RMSE log & \cellcolor{col2}$\delta_1$ & \cellcolor{col2}$\delta_2$ & \cellcolor{col2}$\delta_3$\\
    \hline
    \hline
    HR-Depth~\cite{hrdepth} & 1 & 0.106 & 0.755 & 4.472 & 0.181 & 0.892 & 0.966 & \underline{0.984} \\
    R-MSFM6~\cite{r-msfm} & 1 & 0.108 & 0.748 & 4.470 & 0.185 & 0.889 & 0.963 & 0.982 \\
    BRNet~\cite{brnet} & 1 & 0.103 & 0.684 & 4.385 & 0.175 & 0.889 & 0.965 & \textbf{0.985}\\
    DevNet~\cite{devnet} & 1 & 0.100 & 0.699 & 4.412 & 0.174 & 0.893 & 0.966 & \textbf{0.985}\\
    MUSTNet~\cite{mustnet} & 1 & 0.104 & 0.750 & 4.451 & 0.180 & 0.895 & 0.966 & 0.984 \\
    DepthSegNet~\cite{depthsegnet} & 1 & 0.099 & \underline{0.624} & 4.165 & 0.171 & 0.902 & \underline{0.969} & \textbf{0.985} \\
    DIFFNet~\cite{diffnet} & 1 & 0.097 & 0.722 & 4.435 & 0.174 & \underline{0.907} & 0.967 & \underline{0.984} \\
    SD-SSMDE~\cite{sd-ssmde} & 1 & 0.098 & 0.674 & 4.187 & 0.170 & 0.902 & 0.968 & \textbf{0.985} \\
    DaCCN~\cite{daccn} & 1 & \underline{0.094} & \underline{0.624} & \underline{4.145} & \underline{0.169} & \textbf{0.909} & \textbf{0.970} & \textbf{0.985} \\
    \hdashline  
    Monodepth2 (ResNet18)~\cite{mono2}& 1 & 0.115 & 0.882 & 4.701 & 0.190 & 0.879 & 0.961 & 0.982 \\
    \textbf{Mono-ViFI (ResNet18)}& 1 &0.103 & 0.660 & 4.277 & 0.176 & 0.892 & 0.967 & \textbf{0.985} \\   
    \hdashline 
    Lite-Mono~\cite{lite-mono}& 1  &  0.102 & 0.746 & 4.444 & 0.179 & 0.896 & 0.965 & 0.983 \\
    \textbf{Mono-ViFI (Lite-Mono)}& 1  & 0.101 & 0.655 & 4.229 & 0.173 & 0.897 & 0.967 & \textbf{0.985}\\    
    \hdashline 
    D-HRNet~\cite{radepth} & 1 & 0.100  & 0.715 & 4.304 & 0.176 & 0.903  & 0.967 & \underline{0.984}\\
    \textbf{Mono-ViFI (D-HRNet)} & 1 & \textbf{0.093} & \textbf{0.589} & \textbf{4.072} & \textbf{0.168} & \textbf{0.909} & \underline{0.969} & \textbf{0.985}\\
    \hline 
    ManyDepth (ResNet18)~\cite{manydepth} & 2 (-1, 0) & 0.093 & 0.715 & 4.245 & 0.172 & 0.909 & 0.966 & 0.983\\
    ManyDepth (ResNet50)~\cite{manydepth} & 2 (-1, 0) & \underline{0.091} & 0.694 & 4.245 & 0.171 & \underline{0.911} & 0.968 & 0.983\\
    \textbf{Mono-ViFI (ResNet18)} & 3 (-1, 0, 1) &  0.099 & \underline{0.620} & 4.170 & \underline{0.170} & 0.902 & \underline{0.969} & \underline{0.985}\\
    \textbf{Mono-ViFI (Lite-Mono)} & 3 (-1, 0, 1)& 0.098 & 0.622 & \underline{4.167} & \underline{0.170} & 0.902 & \underline{0.969} & \underline{0.985}\\
    \textbf{Mono-ViFI (D-HRNet)} & 3 (-1, 0, 1) & \textbf{0.089} & \textbf{0.556} & \textbf{3.981} & \textbf{0.164} & \textbf{0.914} & \textbf{0.971} & \textbf{0.986}\\
    \hline 
    \end{tabular}}
    \vspace{-6mm}
    \end{center}
    \end{table*}

\subsection{Scale-invariant Error Function}
To calculate depth consistency loss between two depth maps $D_1$ and $D_2$, we adopt the scale-invariant mean squared error in log space proposed by Eigen \etal~\cite{eigen}, which is widely used in supervised depth estimation. Specifically, the error map between $D_1$ and $D_2$ in log space is formulated as $E=log(D_1)-log(D_2)$. Let $e_i$ denote the $i_{th}$ pixel of $E$, and the scale-invariant error function $\rm SI()$ can be calculated by:
\begin{small}
\begin{equation}
    {\rm SI}(D_1, D_2)=\frac{1}{V} \sum_i e_i^2 - \frac{\beta}{V^2}\left(\sum_i e_i\right)^2,
\end{equation}
\end{small}
where $\beta=0.5$ and $V$ is the number of valid pixels.

\begin{table*}[t]
    \begin{center}
    \caption{Computational efficiency. Enc is encoder. Dec is decoder. SF is single-frame. MF is multi-frame. MACs and runtime are measured under $640\times192$ resolution on one NVIDIA L40S card. Depth metrics are Cityscapes results. All backbones are ResNet18.}
    \vspace{-1mm}
    \label{tab_runtime}
    \resizebox{0.7\textwidth}{!}{
    \begin{tabular}{|c||c|c|c||c|c|c|}
    \hline
    Method & \#Params & MACs & Runtime & \cellcolor{col1}Abs Rel & \cellcolor{col1}RMSE & \cellcolor{col2}$\delta_1$ \\
    \hline
    Enc + Dec (\textbf{Ours-SF}) & 14.3M & 8G & 2.8ms & 0.112 & 5.899 & 0.872 \\
    VFI Model  & 2.8M & 8.1G & 4.4ms & - & - & - \\
    Fusion Module & 0.8M & 0.8G & 6.6ms & - & - & - \\
    \hdashline
    \textbf{Ours-MF} & \textbf{17.9M} & 16.9G & \textbf{13.8ms} & \textbf{0.107} & \textbf{5.688} & \textbf{0.880} \\
    ManyDepth~\cite{manydepth} & 26.9M & \textbf{13.7G} & 16.7ms & 0.114 & 6.223 & 0.875 \\
    \hline
    \end{tabular}}
    \vspace{-2mm}
    \end{center}
    \end{table*}

\begin{table*}[t]
    \begin{center}
        %\vspace{-5mm}
    \caption{Ablation on affine transformation. Scale-aware depth consistency is included.}
    \label{sup:tab2}
    \vspace{-1mm}
    \renewcommand{\arraystretch}{1.1}{
    \resizebox{0.9\textwidth}{!}{
    \begin{tabular}{|c|c||c|c|c|c|c|c|c|c|}
    \hline
    Method & Dataset & \cellcolor{col1}Abs Rel & \cellcolor{col1}Sq Rel & \cellcolor{col1}RMSE & \cellcolor{col1}RMSE log & \cellcolor{col2}$\delta_1$ & \cellcolor{col2}$\delta_2$ & \cellcolor{col2}$\delta_3$ \\
    \hline
    \rowcolor{gray!20} \multicolumn{9}{|c|}{Single-frame Inference} \\
    \hline
    Baseline (Monodepth2) & K & 0.115 & 0.903 & 4.863 & 0.193 & 0.877 & 0.959 & 0.981 \\
    Baseline + resizing-cropping & K &  0.108 &   0.789 &   4.561 &  0.184 &  0.885 &  0.963 &  0.983 \\
    Baseline + rotation & K & 0.110 & 0.787 & 4.637&0.186	&0.882&	0.961&0.982\\
    Baseline + affine transformation & K & \textbf{0.107} & \textbf{0.762} & \textbf{4.461} & \textbf{0.181} & \textbf{0.886} & \textbf{0.964} & \textbf{0.984} \\ 
    \hline
    Baseline (Monodepth2) & CS & 0.129 & 1.569 & 6.876 & 0.187 & 0.849 & 0.957 & 0.983 \\
    Baseline + resizing-cropping & CS & 0.116 & 1.079 & 6.189 & 0.169 & 0.867 & 0.964 & 0.988 \\
    Baseline + rotation & CS & 0.121 & 1.221 & 6.327 & 0.175 & 0.866 & 0.964 & 0.987 \\
    Baseline + affine transformation & CS & \textbf{0.113} & \textbf{1.053} & \textbf{6.050} & \textbf{0.167} & \textbf{0.870} & \textbf{0.966} & \textbf{0.989}  \\ 
    \hline
    \end{tabular}}}
    \vspace{-5mm}
    \end{center}
    \end{table*}

\subsection{Fourier Positional Encoding}
\begin{figure*}[t]
    \centering
    \includegraphics[width=1.0\linewidth]{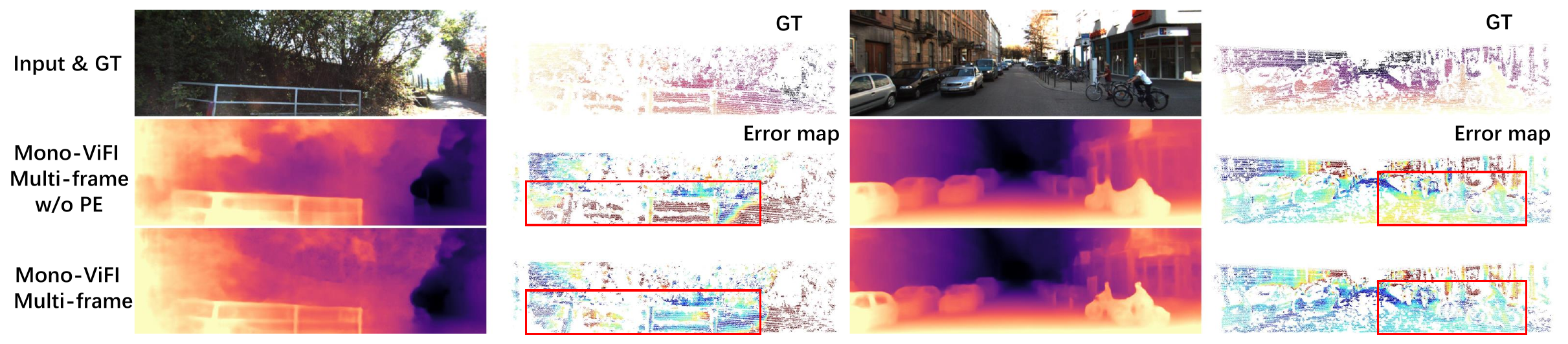}
    \caption{Qualitative ablation results about the fourier positional encoding on KITTI. PE denotes positional encoding. Error maps in columns 2 and 4 show the Abs Rel error compared to the improved ground truth~\cite{im-gt}
    from good (blue) to bad (red).}
    \label{fig:vis_pe}
    \vspace{-1mm}
\end{figure*}

Directly concatenating aligned features with optical flow leads to poor performance (see in \Cref{sup:tab3} and \cref{fig:vis_pe}), which is consistent with the analysis in \cite{nerf,ffl}. Therefore, we follow them to employ a fourier positional encoding function $g : \mathbb{R} \rightarrow \mathbb{R}^{2S+1}$. For an input $u$, we can formulate it as:
\begin{equation}
\begin{aligned}
    g(u) = [u, &~{\rm sin}(2^0 \pi u), ~{\rm cos}(2^0 \pi u), ~\cdots, \\
    &{\rm sin}(2^{S-1} \pi u), ~{\rm cos}(2^{S-1} \pi u)].
\end{aligned}
\end{equation}
The encoding function $g$ is applied separately to each of the two components in a flow field $F=\{F_x, F_y\}$. And we set $S=10$ in our experiments.

\subsection{Additional Explanation for Feature Fusion}
Different to cost volume which explicitly encodes geometry, feature fusion aggregates temporal information to implicitly infer scene depth. First, feature alignment provides spatial relevance among the same location in neighbor frames. Then, injecting pixel motion (optical flow) to feature embeddings implicitly provides relative distance relation. Larger pixel motion usually indicates closer depth. Although this is not suitable for dynamic pixels, the influence is slight since this information is implicit. Finally, the merge mask can alleviate misleading occlusion pixels.

\begin{table*}[t]
\begin{center}
\caption{Quantitative ablation results on fourier positional encoding for optical flow.}
\label{sup:tab3}
\vspace{-1mm}
\renewcommand{\arraystretch}{1.1}{
\resizebox{0.99\textwidth}{!}{
\begin{tabular}{|c|c||c|c|c|c|c|c|c|c|}
\hline
Method & Dataset & \cellcolor{col1}Abs Rel & \cellcolor{col1}Sq Rel & \cellcolor{col1}RMSE & \cellcolor{col1}RMSE log & \cellcolor{col2}$\delta_1$ & \cellcolor{col2}$\delta_2$ & \cellcolor{col2}$\delta_3$ \\
\hline
\rowcolor{gray!20} \multicolumn{9}{|c|}{Multi-frame Inference} \\
\hline
Mono-ViFI w/o fourier positional encoding & K & 0.105 & 0.708 & 4.487 & 0.181 & 0.888 & 0.963 & 0.983\\
Mono-ViFI & K & \textbf{0.099} & \textbf{0.661} & \textbf{4.321} & \textbf{0.174} & \textbf{0.898} & \textbf{0.966} & \textbf{0.985} \\ 
\hline
Mono-ViFI w/o fourier positional encoding & CS & 0.114 & 1.026 & 5.904 & 0.167 & 0.868 & 0.965 & 0.989 \\
Mono-ViFI & CS & \textbf{0.107} & \textbf{0.934} & \textbf{5.688} & \textbf{0.159} & \textbf{0.880} & \textbf{0.969} & \textbf{0.990}  \\ 
\hline
\end{tabular}}}
\vspace{-3mm}
\end{center}
\end{table*}

\section{Training and Evaluation Settings} 
\noindent \textbf{Training.} The depth decoders end with a sigmoid activation function. We follow~\cite{mono2} to convert the sigmoid output $\sigma$ to depth with $D=1/(a\sigma+b)$, where $a$ and $b$ are selected to constrain $D$ between 0.1 and 100 units. In addition to the introduced spatial and temporal augmentation, we also adopt some basic augmentation approaches, including horizontal flipping and the following: random brightness contrast, saturation, and hue jitter. Note that these color augmentations are not applied to the images used to compute the photometric losses. Our training framework is implemented in PyTorch~\cite{pytorch} with AdamW~\cite{adamw} optimizer. Models are trained for $E$ epochs with a learning rate of $10^{-4}$ for the first $E_0$ epochs and $10^{-5}$ for the remainder. For KITTI~\cite{kitti}, we set $E=20$, $E_0=15$, and the resolution is $640 \times 192$ unless otherwise stated. For Cityscapes~\cite{city}, that is $E=10$, $E_0=7$, and a resolution of $512 \times 192$. We additionally train with the high resolution of $1024 \times 320$ for KITTI, where models are initialized using the weights from the medium resolution ($640 \times 192$) models and are then trained for 5 epochs with a learning rate of $10^{-5}$. Different from~\cite{mono2}, we only predict a single depth map at the full scale to save memory and accelerate training. 
All the experiments are conducted on NVIDIA RTX L40S GPUs.

\noindent \textbf{Evaluation.} When evaluating, we cap the depth to 80m on KITTI/Cityscapes and 70m on Make3D~\cite{make3d}. Note that we use the same cropping scheme as ManyDepth~\cite{manydepth} for Cityscapes evaluation, generating a region of $412 \times 128$. And median scaling is employed before evaluation to tackle the scale ambiguity.

\begin{table*}[t]
\begin{center}
    %\vspace{-5mm}
\caption{Ablation results on the scale of VFI model, \ie, IFRNet~\cite{ifrnet}, using the multi-frame model. \#Params and MACs are indexes of IFRNet employed for evaluation. The configuration adopted in our work is colored in \colorbox{orange!25}{orange}. }
\label{sup:tab4}
\vspace{-1mm}
\renewcommand{\arraystretch}{1.1}{
\resizebox{0.9\textwidth}{!}{
\begin{tabular}{|c|c|c|c|c||c|c|c|c|c|c|c|}
\hline
\multicolumn{2}{|c|}{Scale of IFRNet~\cite{ifrnet}} & \multirow{2}{*}{\#Params} & \multirow{2}{*}{MACs} & \multirow{2}{*}{Dataset} & \multicolumn{7}{c|}{Metrics}  \\
\cline{1-2}\cline{6-12}
Train & Test & & & & \cellcolor{col1}Abs Rel & \cellcolor{col1}Sq Rel & \cellcolor{col1}RMSE & \cellcolor{col1}RMSE log & \cellcolor{col2}$\delta_1$ & \cellcolor{col2}$\delta_2$ & \cellcolor{col2}$\delta_3$\\
\hline
\rowcolor{gray!20} \multicolumn{12}{|c|}{Multi-frame Inference} \\
\hline
~~~small~~~ & small  & \textbf{2.8M} & \textbf{8.1G} & K & 0.100 & 0.673 & 4.344 & 0.175 & 0.895 & 0.965 & 0.984\\
\rowcolor{orange!25}
~~~large~~~ & small  & \textbf{2.8M} & \textbf{8.1G} & K & \textbf{0.099} & 0.661 & 4.321 & 0.174 & 0.898 & 0.966 & \textbf{0.985}\\
~~~large~~~ & large  & 19.7M & 54.5G & K & \textbf{0.099} & \textbf{0.655} & \textbf{4.304} & \textbf{0.173} & \textbf{0.899} & \textbf{0.967} & \textbf{0.985}\\
\hline
~~~small~~~ & small & \textbf{2.8M} & \textbf{6.5G} & CS & 0.110 & 0.978 & 5.738 & 0.163 & 0.874 & 0.967 & 0.989\\
\rowcolor{orange!25}
~~~large~~~ & small & \textbf{2.8M} & \textbf{6.5G} & CS & \textbf{0.107} & 0.934 & 5.688 & 0.159 & 0.880 & \textbf{0.969} & \textbf{0.990}\\
~~~large~~~ & large & 19.7M & 43.6G & CS & \textbf{0.107} & \textbf{0.925} & \textbf{5.658} & \textbf{0.158} & \textbf{0.881} & \textbf{0.969} & \textbf{0.990}\\
\hline
\end{tabular}}}
\vspace{-2mm}
\end{center}
\end{table*}
\begin{table*}[t]
\begin{center}
    %\vspace{-5mm}
\caption{Ablation on shared or separate encoders/decoders between the single-frame and multi-frame models. The configuration adopted in our work is colored in \colorbox{orange!25}{orange}. }
\label{sup:tab5}
\vspace{-1mm}
\renewcommand{\arraystretch}{1.1}{
\resizebox{0.85\textwidth}{!}{
\begin{tabular}{|c|c|c||c|c|c|c|c|c|c|}
\hline
\multicolumn{2}{|c|}{Single- and Multi-frame Models}  & \multirow{2}{*}{Dataset} & \multicolumn{7}{c|}{Metrics}  \\
\cline{1-2}\cline{4-10}
Encoder & Decoder & & \cellcolor{col1}Abs Rel & \cellcolor{col1}Sq Rel & \cellcolor{col1}RMSE & \cellcolor{col1}RMSE log & \cellcolor{col2}$\delta_1$ & \cellcolor{col2}$\delta_2$ & \cellcolor{col2}$\delta_3$\\
\hline
\rowcolor{gray!20} \multicolumn{10}{|c|}{Single-frame Inference} \\
\hline
~~~~~~shared~~~~~~ & shared  & K & 0.107 & 0.716 & 4.533 & 0.181 & 0.886 & 0.963 & \textbf{0.984} \\
\rowcolor{orange!25}
~~~shared~~~ & separate & K & \textbf{0.105} & \textbf{0.708} & \textbf{4.446} & \textbf{0.179} & \textbf{0.887} & \textbf{0.965} & \textbf{0.984} \\
~~~separate~~~ & separate  & K & \textbf{0.105} & 0.724 & 4.471 & 0.180 & 0.886 & 0.964 & \textbf{0.984} \\
\hline
~~~shared~~~ & shared & CS & 0.114 & 1.034 & 5.953 & 0.168 & 0.868 & 0.965 & 0.989 \\
\rowcolor{orange!25}
~~~shared~~~ & separate & CS & \textbf{0.112} & \textbf{1.003} & \textbf{5.899} & \textbf{0.165} & 0.872 & \textbf{0.966} & \textbf{0.990} \\
~~~separate~~~ & separate & CS & \textbf{0.112} & 1.026 & 5.912 & 0.166 & \textbf{0.873} & \textbf{0.966} & \textbf{0.990} \\
\hline
\rowcolor{gray!20} \multicolumn{10}{|c|}{Multi-frame Inference} \\
\hline
~~~shared~~~ & shared &  K & 0.100 & 0.655 & 4.311 & 0.175 & \textbf{0.898} & \textbf{0.966} & 0.984 \\
\rowcolor{orange!25}
~~~shared~~~ & separate & K & \textbf{0.099} & 0.661 & 4.321 & \textbf{0.174} & \textbf{0.898} & \textbf{0.966} & \textbf{0.985} \\
~~~separate~~~ & separate & K & \textbf{0.099} & \textbf{0.652} & \textbf{4.300} & \textbf{0.174} & \textbf{0.898} & \textbf{0.966} & \textbf{0.985} \\
\hline
~~~shared~~~ & shared & CS & 0.111 & 0.972 & 5.732 & 0.162 & 0.874 & 0.967 & \textbf{0.990} \\
\rowcolor{orange!25}
~~~shared~~~ & separate &  CS & \textbf{0.107} & \textbf{0.934} & \textbf{5.688}& \textbf{0.159} & 0.880 & \textbf{0.969} & \textbf{0.990} \\
~~~separate~~~ & separate & CS & \textbf{0.107} & 0.949 & 5.692 & \textbf{0.159} & \textbf{0.881} & \textbf{0.969} & \textbf{0.990} \\
\hline
\end{tabular}}}
\vspace{-4mm}
\end{center}
\end{table*}

\section{Additional Experimental Results}
\subsection{Quantitative Performance Comparison}
As supplementary, we compare proposed Mono-ViFI with other methods on KITTI benchmark with a high resolution of $1024 \times 320$. Results are shown in \Cref{sup:tab1}, which again demonstrate the effectiveness of our approach.

\subsection{Computational Efficiency and Runtime} We have presented model parameters and multiply-add computations (MACs) in Table 1 of main paper. Here we list more detailed results in \Cref{tab_runtime}. We give the average MACs and runtime per frame when processing a long video sequence, since extracted multi-frame features can be reused in a video. The runtime of VFI model and multi-frame fusion module is 2-3 times of single-frame runtime. For fusion module, it has few network parameters and MACs (only $1 \times 1$ convolutions). Most part of its runtime comes from feature warping and fourier positional encoding (\textit{sin/cos}) of optical flow. Even though, the speed of our multi-frame model is still faster than ManyDepth~\cite{manydepth}.

\begin{figure*}[t]
    \centering
    \includegraphics[width=1.0\linewidth]{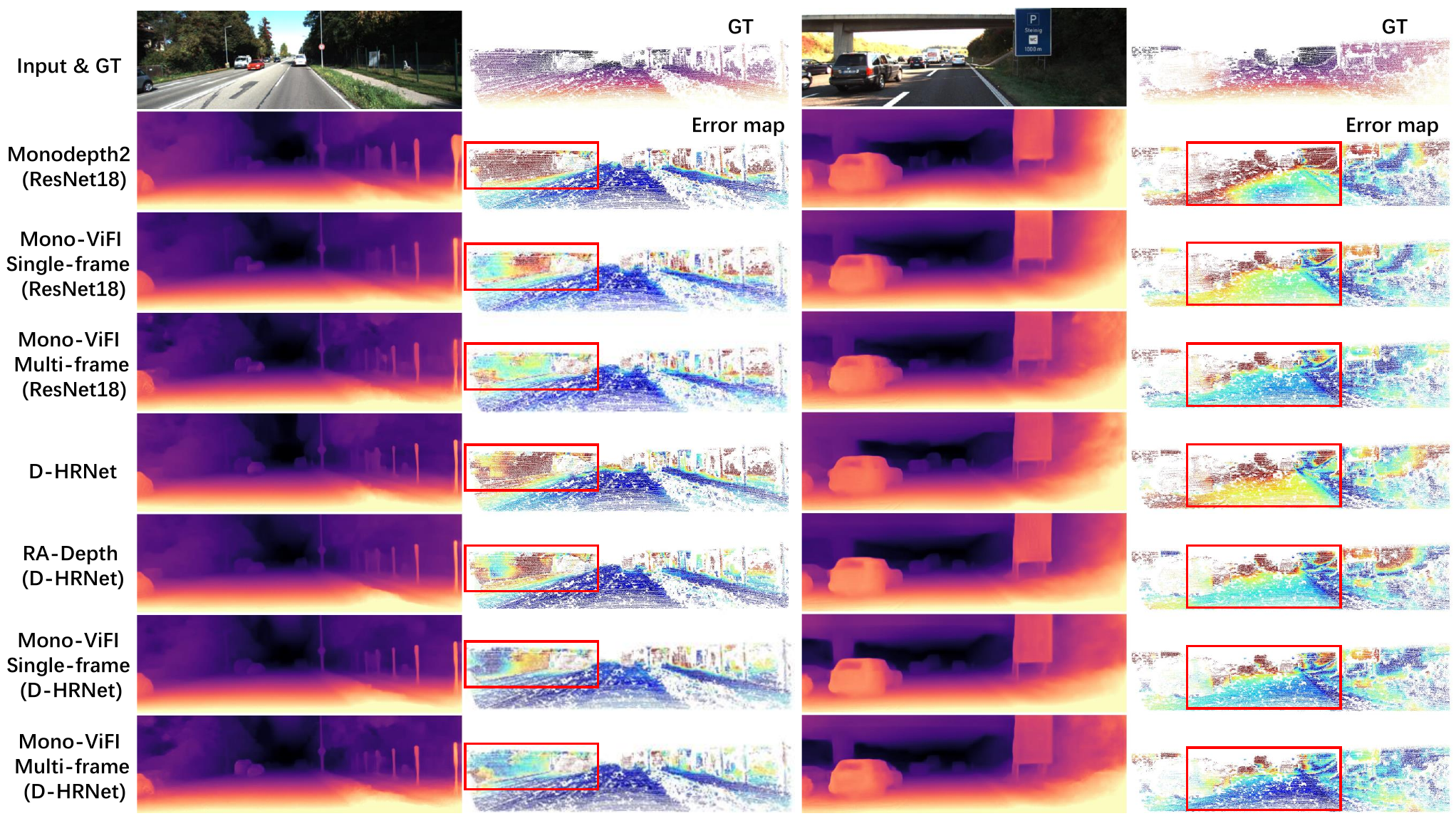}
    \caption{More visualization examples on KITTI~\cite{kitti}. Error maps in columns 2 and 4 show the Abs Rel error compared to the improved ground truth~\cite{im-gt} from good (blue) to bad (red). All error maps are colormapped equivalently.}
    \label{fig:vis_kitti}
    \vspace{-3mm}
\end{figure*}

\subsection{Ablation Study}
In this section, we provide more ablation analysis of Mono-ViFI. Ablation experiments are conducted on both KITTI (K) and Cityscapes (CS) datasets with ResNet18 backbone. For KITTI, raw ground truth is used when evaluating.

\noindent \textbf{Affine Transformation.} The detailed ablation results about our spatial augmentation, \ie, the affine transformation, are listed in \Cref{sup:tab2}. The effect of resizing-cropping has already been revealed in~\cite{radepth,bdedepth}. We show that image rotation can also make sense by creating different scene structures. Moreover, when combining resizing-cropping and rotation into affine transformation, the model performance can be further improved. 

\noindent \textbf{Fourier Positional Encoding for Optical Flow.} To observe the influence of fourier positional encoding for optical flow, we give the quantitative and qualitative ablation results in \Cref{sup:tab3} and \cref{fig:vis_pe}, respectively. Obviously, there is a significant performance decline if it is removed.

\noindent \textbf{Scale of the VFI Model.} The VFI process acts as a module of our multi-frame Mono-ViFI. Consequently, the scale of the VFI model, IFRNet~\cite{ifrnet} can affect the model performance. We mainly experiment with large and small IFRNets, as show in \Cref{sup:tab4}. It can be seen that large IFRNet performs better than small one, but with more parameters and multiply-add computations (MACs). Therefore, we train with large IFRNet to fully leverage its capability while evaluating with small to reduce inference complexity. This configuration is just slightly worse than employing the large model for both training and evaluation.

\begin{figure*}[t]
    \centering
    \includegraphics[width=1.0\linewidth]{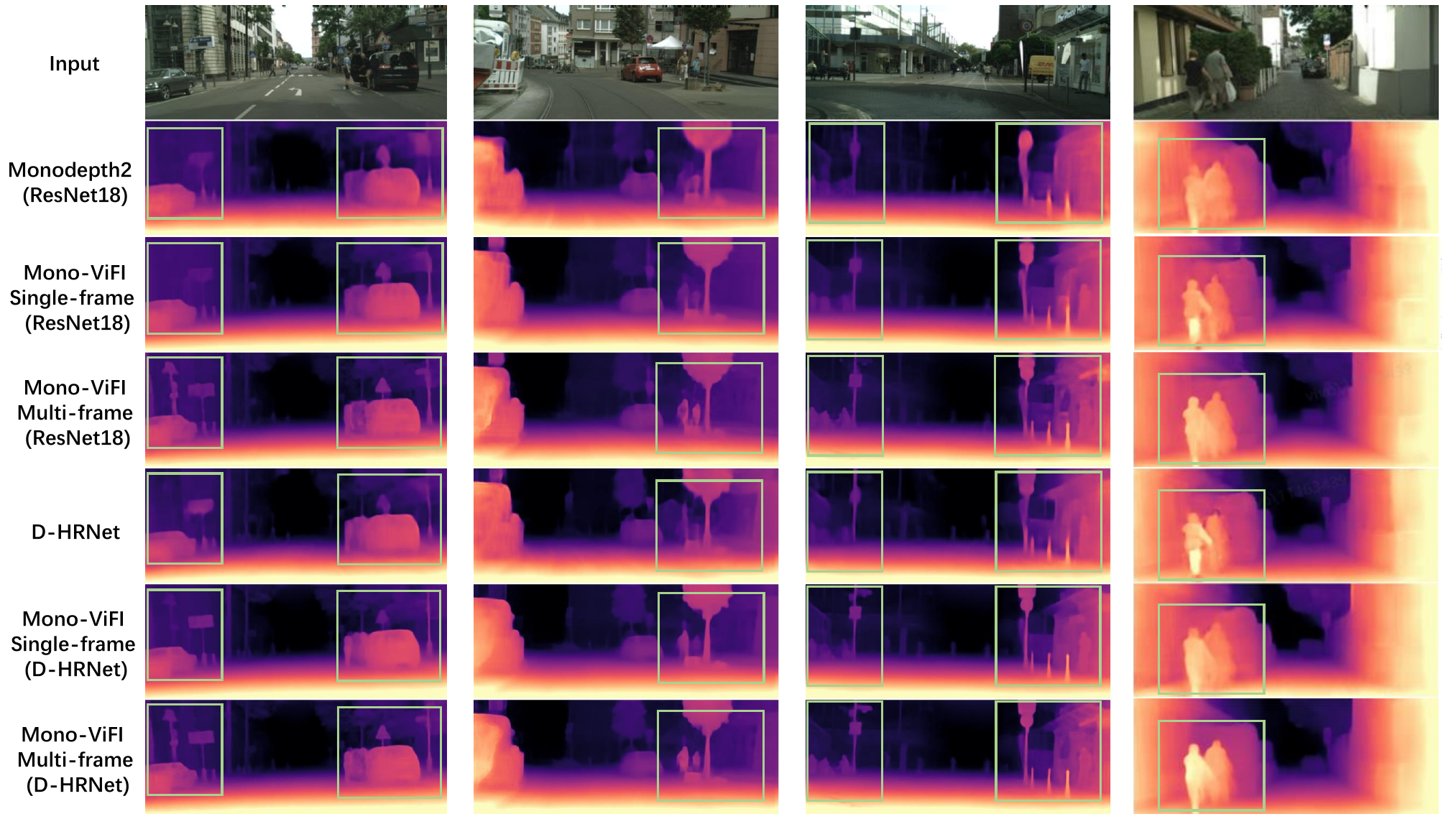}
    \caption{Visualization examples on Cityscapes~\cite{city}.}
    \label{fig:vis_cs}
\end{figure*}

\begin{figure}[t]
    \centering
    \includegraphics[width=0.5\linewidth]{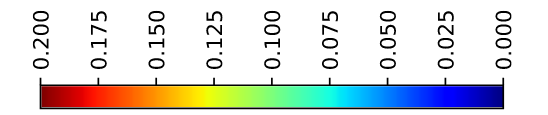}
    \caption{Color scale used for all error plots in the main paper and supplementary material. Values on the color axis are units of Abs Rel error, which is computed after conducting median scaling for the predicted depth.}
    \label{fig:colorbar}
    \vspace{-7mm}
\end{figure}

\noindent \textbf{Shared or Separate Encoders/Decoders.} Note that in Mono-ViFI, the single- and multi-frame models share the same encoder but with separate decoders. Even so, their decoders adopt the same architecture. Here we additionally explore other possibilities, including sharing both the encoder and decoder or completely separating them, as depicted in \Cref{sup:tab5}. Comparing to the configuration employed in Mono-ViFI, sharing both encoder and decoder can result in a performance drop, which is more noticeable on Cityscapes. This is mainly due to the decoding conflicts between single-and multi-frame models. Besides, the performance under fully separating closely mirrors that of our configuration, which means that the same features can be applied to both single- and multi-frame reasoning. To simplify the overall framework and reduce the memory cost, we choose to share the encoder.

\subsection{More Visualizations}
\subsubsection{KITTI and Cityscapes.}
In this section, more visualization results are provided to qualitatively compare Mono-ViFI with other methods. We mainly use two networks as baseline models, Monodepth2-ResNet18~\cite{mono2} and D-HRNet~\cite{radepth}. The KITTI visualization examples are shown in \cref{fig:vis_kitti} and  Cityscapes examples are in \cref{fig:vis_cs}. Similarly to~\cite{manydepth}, we present the error maps for KITTI using the improved ground truth~\cite{im-gt}, which depicts the absolute relative (Abs Rel) error in depth, computed by:
\begin{equation}
    {\rm Abs~Rel} = \frac{|D_{\rm pred}-D_{\rm gt}|}{D_{\rm gt}}.
\end{equation}
The colorbar for displaying error maps is shown in \cref{fig:colorbar}.

\begin{figure*}[t]
    \centering
    \includegraphics[width=0.9\linewidth]{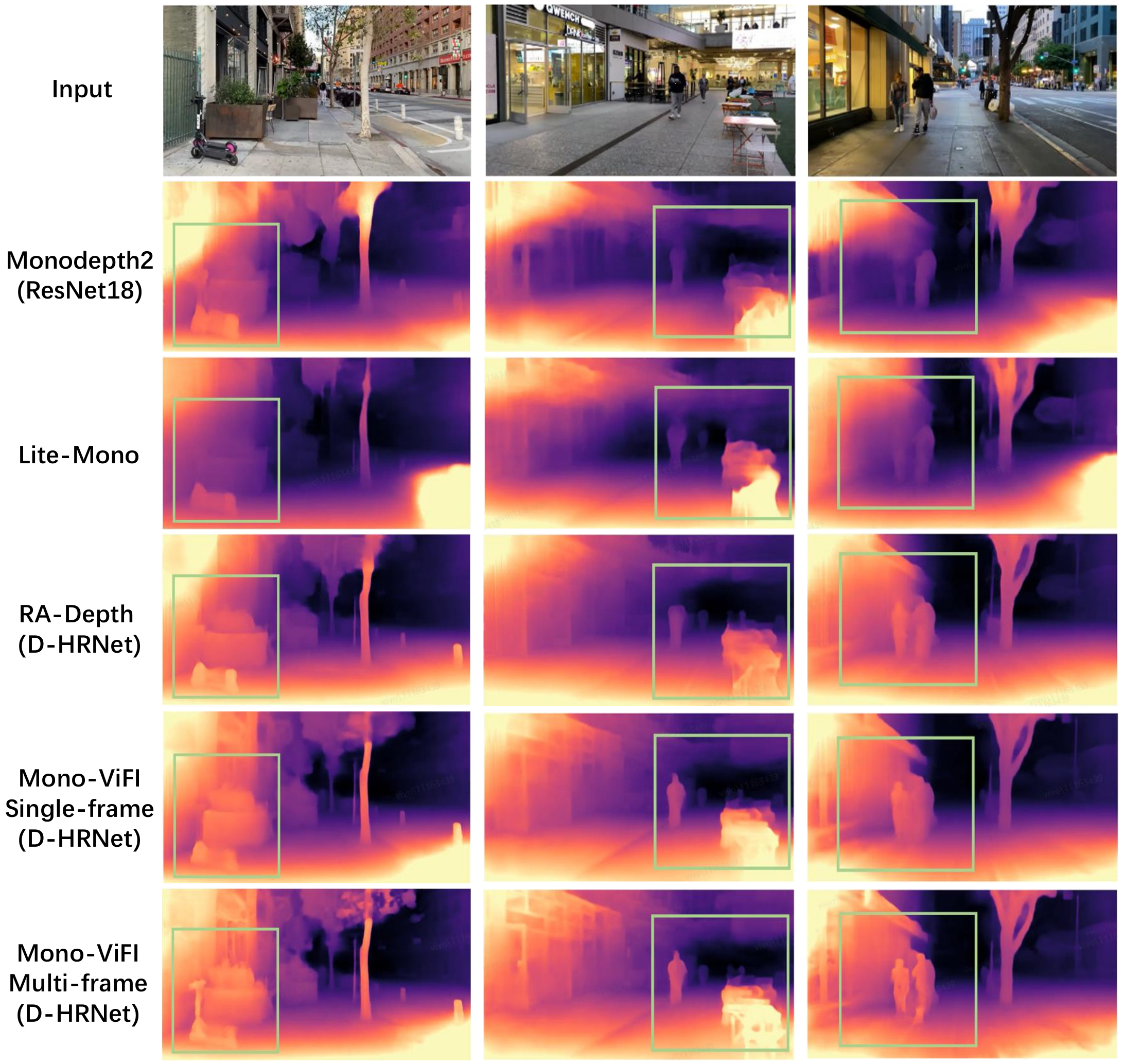}
    \caption{More qualitative results on YouTube video frames. All depth maps are predicted under the resolution of $640 \times 360$ using the corresponding models, which are trained on the KITTI~\cite{kitti} dataset with a resolution of $640 \times 192$. }
    \label{fig:vis_youtube_sup}
\end{figure*}

\subsubsection{YouTube Video}
To qualitatively demonstrate the generalization ability of Mono-ViFI, we use the KITTI-trained models to predict the depths for frames from a YouTube video, downloaded from the Wind Walk Travel Videos channel. The KITTI training resolution is $640 \times 192$, while the evaluation resolution is $640 \times 360$. One example has already been presented in the main paper. Here we give more visualization results, as illustrated in \cref{fig:vis_youtube_sup}, which show that our method can obtain higher quality depth maps with finer details when transferring between different datasets.

\bibliographystyle{splncs04}
\bibliography{main}
\end{document}